%% file: main.tex
\documentclass{article}
\usepackage[accepted]{icml2019}

\usepackage{url}
\usepackage{booktabs}
\usepackage{amsfonts}
\usepackage{amssymb}
\usepackage{nicefrac}
\usepackage{microtype}
\usepackage{amsmath}
\usepackage{subcaption}
\usepackage{dblfloatfix}
\usepackage[title]{appendix}
\usepackage{ragged2e}
\usepackage[labelfont=bf,justification=justified,format=plain]{caption}
\usepackage{colortbl}
\usepackage{multirow}

\usepackage[symbol]{footmisc}

\usepackage[symbol]{footmisc}

\usepackage{microtype}
\usepackage{graphicx}
\usepackage{booktabs} 

\makeatletter
\renewcommand{\paragraph}{%
  \@startsection{paragraph}{4}%
  {\z@}{0.25ex \@plus 0.25ex \@minus .2ex}{-1em}%
  {\normalfont\normalsize\bfseries}%
}
\makeatother

\input{math_commands.tex}

\newenvironment{enum}{
\begin{enumerate}
 \vspace{-3mm}
  \setlength{\leftskip}{-3mm}
  \setlength{\itemsep}{0pt}
  \setlength{\parskip}{0pt}
  \setlength{\parsep}{0pt}
}{\end{enumerate}}
\newenvironment{ls}{
\begin{enumerate}
  \vspace{-3mm}
  \setlength{\leftskip}{-3mm}
  \setlength{\itemsep}{0pt}
  \setlength{\parskip}{0pt}
  \setlength{\parsep}{0pt}
}{\end{enumerate}}

\icmltitlerunning{Dynamic sparse reparameterization}

\begin{document}

\twocolumn[
\icmltitle{Parameter Efficient Training of Deep Convolutional \\ Neural Networks by Dynamic Sparse Reparameterization}

\icmlsetsymbol{equal}{*}

\begin{icmlauthorlist}
\icmlauthor{Hesham Mostafa}{int}
\icmlauthor{Xin Wang}{int,ext}
\end{icmlauthorlist}

\icmlaffiliation{int}{Artificial Intelligence Products Group, Intel Corporation, San Diego, CA, USA.}
\icmlaffiliation{ext}{Currently with Cerebras Systems, Los Altos, CA, USA, work done while at Intel Corporation}

\icmlcorrespondingauthor{Xin Wang}{xin@cerebras.net}

\icmlkeywords{sparse, reparameterization, convolutional neural network, compression, pruning}

\vskip 0.3in
]

\printAffiliationsAndNotice{}  

\input{./abstract.tex}
\input{./introduction.tex}
\input{./related_work.tex}
\input{./methods.tex}

\input{./experimental_results.tex}

\input{./discussion.tex}


\clearpage
\input{./supplementary.tex}

\end{document}

%% file: math_commands.tex
\usepackage{amsmath,amsfonts,bm}
\usepackage{mathtools}

\def\eqref#1{equation~\ref{#1}}

\def\1{\bm{1}}

\def\vzero{{\bm{0}}}

\def\vtheta{{\bm{\theta}}}
\def\vphi{{\bm{\phi}}}
\def\vpsi{{\bm{\psi}}}

\def\vx{{\bm{x}}}
\def\vy{{\bm{y}}}

\DeclareMathAlphabet{\mathsfit}{\encodingdefault}{\sfdefault}{m}{sl}
\SetMathAlphabet{\mathsfit}{bold}{\encodingdefault}{\sfdefault}{bx}{n}
\newcommand{\tens}[1]{\bm{\mathsfit{#1}}}

\def\tW{{\tens{W}}}

\DeclarePairedDelimiter\abs{\big\lvert}{\big\rvert}
\DeclarePairedDelimiter\round{\big\lfloor}{\big\rceil}
\DeclarePairedDelimiter\Round{\Big\lfloor}{\Big\rceil}
\DeclarePairedDelimiter\ROUND{\Bigg\lfloor}{\Bigg\rceil}

%% file: abstract.tex
\begin{abstract}
  
Modern deep neural networks are typically highly overparameterized.
Pruning techniques are able to remove a significant fraction of network parameters with little loss in accuracy. 
Recently, techniques based on dynamic reallocation of non-zero parameters have emerged, allowing direct training of sparse networks without having to pre-train a large dense model. 
Here we present a novel dynamic sparse reparameterization method that addresses the limitations of previous techniques such as high computational cost and the need for manual configuration of the number of free parameters allocated to each layer. 
We evaluate the performance of dynamic reallocation methods in training deep convolutional networks and show that our method outperforms previous static and dynamic reparameterization methods, yielding the best accuracy for a fixed parameter budget, on par with accuracies obtained by iteratively pruning a pre-trained dense model. 
We further investigated the mechanisms underlying the superior generalization performance of the resultant sparse networks. 
We found that neither the structure, nor the initialization of the non-zero parameters were sufficient to explain the superior performance. 
Rather, effective learning crucially depended on the continuous exploration of the sparse network structure space during training. 
Our work suggests that exploring structural degrees of freedom during training is more effective than adding extra parameters to the network.

\end{abstract}

%% file: introduction.tex
\section{Introduction}

Deep neural networks' success in a wide range of application domains, ranging from computer vision to machine translation to automatic speech recognition, stems from their ability to learn complex transformations by data examples while achieving superior generalization performance. 
Though they generalize well, deep networks learn more effectively when they are highly overparameterized~\citep{Brutzkus2017,Zhang2016}. 
Emerging evidence has attributed this need for overparameterization to the geometry of the high-dimensional loss landscapes~\citep{Dauphin2014,Choromanska2014,Goodfellow2014,Im2016,Wu2017,Liao2017,Cooper2018,Novak2018}, and to the implicit regularization properties of stochastic gradient descent (SGD)~\citep{Brutzkus2017,Zhang2018,Poggio2018a}, though a thorough theoretical understanding is not yet complete.

In practice, multiple techniques are able to compress large trained models, including distillation~\citep{Bucilua2006,Hinton2015}, weight precision reduction~\citep{Hubara2016,McDonnell2018}, low-rank decomposition~\citep{Jaderberg2014,Denil2013}, and pruning~\citep{Han2015a,Zhang2018b}. 
While these methods are highly effective in reducing the size of network parameters with little degradation in accuracy, they either operate on a pre-trained model or require the full overparameterized model to be stored and updated during, or at least at a certain stage of, training. 
Thus, training remains memory-inefficient despite the compact size of the resultant network produced by compression. 
The effectiveness of these compression methods, however, indicates the existence of compact network parameter configurations that are able to generalize on par with large networks. 
This raises a tantalizing hypothesis that overparameterization during training might not be a strict necessity and alternative training or reparameterization methods might exist to discover and train compact networks directly.

The problem of achieving training-time parameter efficiency is being approached in a number of ways.  
Most straightforward is to search for more parameter efficient network architectures.  
Innovations in this direction for deep convolutional neural networks (CNNs) include adoption of skip connections~\cite{He2015}, replacement of fully-connected layers with global average pooling layers followed directly by the classifier layer ~\cite{Lin}, and depthwise separable convolutions~\cite{sifre2014rigid,Howard2017}. 
These modern CNN architectures drastically improved the accuracies achievable at a given parameter budget. 

Instead of inventing new network architectures, an alternative approach is to reparameterize an existing model architecture, which is the approach we take in this work.  
In general, any \emph{differentiable reparameterization} can be used to augment training of a given model.  
Let an original network (or a layer therein) be denoted by $\vy = f(\vx; \vtheta)$, parameterized by $\vtheta  \in \mathbf{\Theta}$. 
Reparameterize it by $\vphi \in \mathbf{\Phi}$ and $\vpsi \in \mathbf{\Psi}$ through $\vtheta = g(\vphi; \vpsi)$, where $g$ is differentiable w.r.t. $\vphi$ but not necessarily w.r.t. $\vpsi$.  
Denote the reparameterized network by $\displaystyle f_\vpsi$, considering $ \vpsi$ as \emph{metaparameters}~\footnote{
  We use the term \emph{metaparameter} to refer to the parameters $\vpsi$ of the reparameterization function $g$.  
  They differ from parameters $\vphi$ in that they are not optimized through gradient descent, and they differ from hyperparameters in that they define meaningful features of the model which are required for inference. 
}: 
\begin{align}
  \vy =  f\left( \vx; g(\vphi; \vpsi) \right) \triangleq f_\vpsi\left(\vx; \vphi\right) .
\end{align}
$f_\vpsi$ is trained by backpropagating errors through $g$, as $\frac \partial {\partial \vphi} = \frac {\partial g} {\partial \vphi} \frac \partial {\partial g}$. 
If it is so chosen that $\dim(\mathbf{\Phi}) + \dim(\mathbf{\Psi}) < \dim(\mathbf{\Theta})$ and $f_\vpsi \approx f$ in terms of generalization performance, then $f_\vpsi$ is a more parameter efficient function approximator than  $f$.

\emph{Sparse reparameterization} is a special case where $g$ is a linear projection; $\vphi$ is the non-zero entries (i.e. ``weights") and $\vpsi$ their indices (i.e. ``connectivity") in the original parameter tensor $\vtheta$. 
Likewise, \emph{parameter sharing} is a similar special case of linear reparameterization where $\vphi$ is the tied parameters and $\vpsi$ the indices at which each parameter is placed (with repetition) in the original parameter tensor $\vtheta$. 
If metaparameters $\vpsi$ are fixed during the course of training, the reparameterization is \emph{static}, whereas if $\vpsi$ is adjusted adaptively during training, we call it \emph{dynamic} reparameterization.

In this paper, we investigate multiple static and dynamic reparameterizations of deep residual CNNs for efficient training.  
Inspired by previous techniques, we developed a novel dynamic reparameterization method that yielded the highest parameter efficiency in training sparse deep residual networks, outperforming existing static and dynamic reparameterization methods. 

Our method dynamically changes the sparse structure of the network during training.
Its superior performance suggests that, given a limited storage and computational budget for training a CNN, it is better to allocate part of the resources to describing and evolving the structure of the network, than to spend it entirely on the parameters of a dense network.

Furthermore, we show that the success of dynamic sparse reparameterization is not solely due to the final sparse structure of the resultant networks, nor to a combination of final structure and initial weight values. 
Rather, training-time structural exploration is necessary for best generalization, even if a high-performance structure and its initial values are known \textit{a priori}. 
Thus, simultaneous exploration of network structure and parameter optimization through gradient descent are synergistic.
Structural exploration improves the trainability of sparse deep CNNs.

%% file: related_work.tex
\section{Related work}
\label{sec:rel_work}

Training of differentiably reparameterized networks has been proposed in numerous studies before.  

\paragraph{Dense reparameterization}
Several dense reparameterization techniques sought to reduce the size of fully connected layers. 
These include low-rank decomposition~\citep{Denil2013}, fastfood transform~\citep{Yang2014}, ACDC transform~\citep{Moczulski2015}, HashedNet~\citep{Chen2015a}, low displacement rank~\citep{Sindhwani2015} and block-circulant matrix parameterization~\citep{Treister2018}. 

Note that similar reparameterizations were also used to introduce certain algebraic properties to the parameters for purposes other than reducing model sizes, e.g. to make training more stable as in unitary evolution RNNs~\citep{Arjovsky2015} and in weight normalization~\citep{Salimans2016}, to inject inductive biases~\citep{Thomas2018}, and to alter~\citep{Dinh2017} or to measure~\citep{Li2018} properties of the loss landscape. 
These dense reparameterization methods are static.

\paragraph{Sparse reparameterization}
Successful training of sparse reparameterized networks usually employs iterative pruning and retraining, e.g. \cite{Han2015,Narang2017,Zhu2017}~\footnote{
  Note that these, as well as all other sparse techniques we benchmark against in this paper, impose \emph{non-structured} sparsification on parameter tensors, yielding \emph{sparse} models. 
  There also exist a class of \emph{structured} pruning methods that ``sparsify'' at channel or layer granularity, e.g.~\cite{Luo2017} and \cite{Huang2017}, generating essentially small \emph{dense} models.
  We describe a full landscape of existing methods in Appendix~D.
}.  
Training typically starts with a large pre-trained model and sparsity is gradually increased by pruning and fine-tuning.  
Training a small, static, and sparse model \textit{de novo} fares worse than compressing a large dense model to the same sparsity~\citep{Zhu2017}.  

\cite{Frankle2018} identified small and sparse subnetworks post-training which, when trained in isolation, reached a similar accuracy as the enclosing big network. 
They further showed that these subnetworks were sensitive to initialization, and hypothesized that the role of overparameterization is to provide a large number of candidate subnetworks, thereby increasing the likelihood that one of these subnetworks will have the necessary structure and initialization needed for effective learning. 

Most closely related to our work are dynamic sparse reparameterization techniques that emerged only recently.  
Like ours, these methods adaptively alter, by certain heuristic rules, the location of non-zero parameters during training.  
Sparse evolutionary training (SET)~\citep{Mocanu2018} used magnitude-based pruning and random growth at the end of each training epoch. 
NeST~\citep{Dai2017,Dai2018} iteratively grew and pruned parameters and neurons during training; parameter growth was guided by parameter gradient and pruning by parameter magnitude. 
Deep Rewiring (DeepR)~\citep{Bellec2017} combined dynamic sparse parameterization with stochastic parameter updates for training. 
These methods were primarily demonstrated with sparsification of fully-connected layers and applied to relatively small and shallow networks.
They also required manual configuration of sparsity levels for different layers of the model. 
The method we propose in this paper is more scalable and computationally efficient than these previous approaches, while achieving better performance on deep CNNs. 

%% file: methods.tex
\section{Methods}
\label{sec:meth}

\begin{algorithm*}[t]
  \caption{Reallocation of non-zero parameters within and across parameter tensors}
  \label{algo:main}
  \begin{algorithmic}[1]
    \For {each sparse parameter tensor ${\tW}_i$}
    \State $({\tW}_i,k_i) \gets \texttt{prune\_by\_threshold}({\tW}_i,H)$ \Comment{$k_i$ is the number of pruned weights}
    \State $l_i  \gets \texttt{number\_of\_nonzero\_entries}({\tW}_i)$ \Comment{Number of surviving weights after pruning}
    \EndFor
    \State $\left(K, L\right) \gets \left(\sum_i k_i, \sum_i l_i \right)$ \Comment{Total number of pruned and surviving weights}
    \State $H \gets \texttt{adjust\_pruning\_threshold}(H, K, \delta)$ \Comment{Adjust pruning threshold}
    \For {each sparse parameter tensor ${\tW}_i$}
    \State ${\tW}_i \gets \texttt{grow\_back}({\tW}_i,\frac{l_i}{L}K)$ \Comment{Grow $\frac{l_i}{L}K$ zero-initialized weights at random in ${\tW}_i$ }
    \EndFor
  \end{algorithmic}
\end{algorithm*}

We sparsely reparameterize the majority of layers in deep CNNs.  
All sparse parameter tensors are randomly initialized at the same sparsity (i.e. fraction of zeros).
During training, free parameters are moved within and across weight tensors every few hundred training iterations, following a two-phase procedure (Algorithm~\ref{algo:main}): magnitude-based pruning followed by random growth. 
Throughout training, we always maintain the same total number of non-zero parameters in the network. 

Our magnitude-based pruning is based on an adaptive global threshold $H$ where all sparse weights with absolute values smaller than $H$ are pruned. 
$H$ is adjusted via a setpoint negative feedback loop to maintain approximately (with tolerance $\delta$) a fixed number of pruned/grown parameters $N_p$ during each reallocation step. 

Immediately after removing $K$ parameters during the pruning phase, $K$ zero-initialized parameters are redistributed among the sparse parameter tensors, following a heuristic rule: layers having larger fractions of non-zero weights receive proportionally more free parameters (see Algorithm~\ref{algo:main}). 
Intuitively, one should allocate more parameters to layers such that training loss is more quickly reduced. 
This means, to the first order, free parameters should be redistributed to layers whose parameters receive larger loss gradients. 
If a layer has been heavily pruned, this indicates that, for a large portion of its parameters, the training loss gradients are not large or consistent enough to counteract the pull towards zero exerted by weight-decay regularization. 
This layer is therefore to receive a smaller share of new free parameters during the growth phase.
The reallocated parameters are randomly placed at zero positions in the target weight tensors. 
To ensure the numbers of pruned and grown free parameters are exactly the same, we impose extra guards against rounding errors, as well as against special cases where more free parameters are allocated to a tensor than there are zero positions.  
For simplicity of exposition, we omit these minor details in Algorithm~\ref{algo:main}.  
See Appendix~A for a more detailed description of the algorithm.

Our algorithm differs from \emph{SET}~\cite{Mocanu2018} in two important aspects.
First, instead of pruning a fixed fraction of weights at each reallocation step, we use an adaptive threshold for pruning. 
Second, we automatically reallocate parameters across layers during training and do not impose a fixed, manually configured, sparsity level on each layer. 
The first difference leads to reduced computational overhead as it obviates the need for sorting operations, and the second to better performing networks (see Section~\ref{sec:res}) and the ability to train extremely sparse networks as shown in Appendix~F. 

We evaluated our method on the deep residual CNNs listed in Table~\ref{tb:exp}, and compared its performance against existing static and dynamic reparameterization methods\footnote{Code is available at \url{https://github.com/IntelAI/dynamic-reparameterization}.}.
We did not experiment with AlexNet~\citep{Krizhevsky2012} and VGG-style networks~\citep{Simonyan2014} as their parameter efficiency is inferior to modern residual networks.   
Such a choice makes the improvement in parameter efficiency achieved by our dynamic sparse training method more practically relevant. 
Dynamic sparse reparameterization was applied to all weight tensors of convolutional layers (with the exception of downsampling convolutions and the first convolutional layer acting on the input image), while all biases and parameters of normalization layers were kept dense. 

At a specific global sparsity\footnote{
 Global sparsity $s$ is defined as the overall sparsity of all sparse parameter tensors, not the entire model, which has a small fraction of dense parameters.
} $s$, we compared our method (\emph{dynamic sparse}) against six baselines:
\begin{ls}
  \item \emph{Full dense}: original large and dense model, with $N$ free parameters;
  \item \emph{Thin dense}: original model with less wide layers, such that it had the same size as \emph{dynamic sparse};
  \item \emph{Static sparse}: original model initialized at sparsity level $s$ with random sparseness pattern, then trained with connectivity (sparseness pattern) fixed;
  \item \emph{Compressed sparse}: sparse model obtained by iteratively pruning and re-retraining a large and dense pre-trained model to target sparsity $s$~\citep{Zhu2017};
  \item \emph{DeepR}: sparse model trained by using Deep Rewiring~\citep{Bellec2017}; 
  \item \emph{SET}: sparse model trained by using Sparse Evolutionary Training (SET)~\citep{Mocanu2018}. 
\end{ls}
Appendix~C compares our method against an additional static (dense) parameterization method based on weight sharing: \emph{HashedNet}~\citep{Chen2015a}.

Because sparse tensors require storage of both the free parameter values and their locations, we compare models that have the same size in descriptive length, instead of the same number of weights. 
While the number of bits needed to specify the connectivity is implementation dependent, we assume one bit is used for each position in the weight tensors indicating whether this position is zero or not. 
A sparse tensor is fully defined by this bit-mask, together with the non-zero entries. 
This scheme was previously used in CNN accelerators that natively operate on sparse structures~\cite{aimar2018nullhop}. 
For a network with $N$ 32-bit weights in its dense form, a sparse version at sparsity $s$ has a descriptive length of $(32s+1)N$ bits, and is thus equivalent in size to a thinner dense network with $(s+\frac{1}{32})N$ weights. 
We use this formula to determine the parameter counts of the \emph{thin dense} baseline, which has $\frac {N}{32}$ more weights than comparable sparse models. 

A recent study~\cite{Liu2018} showed that training small networks \textit{de novo} can almost always match the generalization performance obtained by post-training pruning of larger networks, so long as the small networks were trained for long enough. 
To address concerns that the superior performance of \emph{dynamic sparse} might be matched by training \emph{thin dense} or \emph{static sparse} networks for more epochs, we always train \emph{thin dense} and \emph{static sparse} baselines for double the number of epochs used to train \emph{dynamic sparse} models. 

Note that \emph{compressed sparse} is a compression method that first trains a large dense model and then iteratively prunes it down, whereas \emph{dynamic sparse} and all other baselines maintain a constant small model size throughout training. 
For \emph{compressed sparse}, we train the large dense model for the same number of epochs used for our \emph{dynamic sparse}, and then iteratively prune and fine-tune it across additional training epochs. 
\emph{Compressed sparse} thus trains for more epochs than \emph{dynamic sparse}. 
See Appendix~B for hyperparameters used for all experiments.

\begin{table*}[t]
\caption{
  Datasets and models used in experiments 
}
\label{tb:exp}
\vspace{-3mm}\centering
\setlength\tabcolsep{5pt}
\begin{tabular}{ l | l | l }
  \toprule
  Dataset & CIFAR10 & Imagenet \\ \midrule
  Model 
    & \begin{tabular}[t]{@{}l}WRN-28-2  { \citep{Zagoruyko2016}}\end{tabular} 
    & \begin{tabular}[t]{@{}l}Resnet-50  { \citep{He2015}}\end{tabular} \\ \midrule
  Architecture 
    & \begin{tabular}[t]{@{}l@{}l@{}l@{}l@{}}
        { C16/3$\times$3} \\ 
        { [C16/3$\times$3,C16/3$\times$3]$\times$4} \\ 
        { [C64/3$\times$3,C64/3$\times$3]$\times$4} \\ 
        { [C128/3$\times$3,C128/3$\times$3]$\times$4} \\ 
        { GlobalAvgPool,\,F10} 
      \end{tabular}  
    & \begin{tabular}[t]{@{}l@{}l@{}l@{}l@{}l@{}}
        { C64/7$\times$7-2,\,MaxPool/3$\times$3-2} \\ 
        { [C64/1$\times$1,\,C64/3$\times$3,\,C256/1$\times$1]$\times$3} \\ 
        { [C128/1$\times$1,\,C128/3$\times$3,\,C512/1$\times$1]$\times$4} \\ 
        { [C256/1$\times$1,\,C256/3$\times$3,\,C1024/1$\times$1]$\times$6} \\ 
        { [C512/1$\times$1,\,C512/3$\times$3,\,C2048/1$\times$1]$\times$3} \\ 
        { GlobalAvgPool,\,F1000} 
      \end{tabular} \\ \midrule  
  \# Parameters  & 1.5M  & 25.6M \\ 
  \bottomrule
\end{tabular}  
\captionsetup{singlelinecheck=off,font=footnotesize,width=0.89\textwidth}
\caption*{
  For brevity architecture specifications omit batch normalization and activations. Pre-activation batch normalization was used in all cases.  
  Convolutional (C) layers are specified with output size and kernel size and Max pooling (MaxPool) layers with kernel size. 
  Brackets enclose residual blocks postfixed with repetition numbers; the downsampling convolution in the first block of a scale group is implied.
}
\end{table*}

%% file: experimental_results.tex
\section{Experimental results}
\label{sec:res}

\begin{figure}[h]
  \centering
  \begin{subfigure}[t]{0.45\textwidth}
    \phantomsubcaption
    \includegraphics[width=\textwidth]{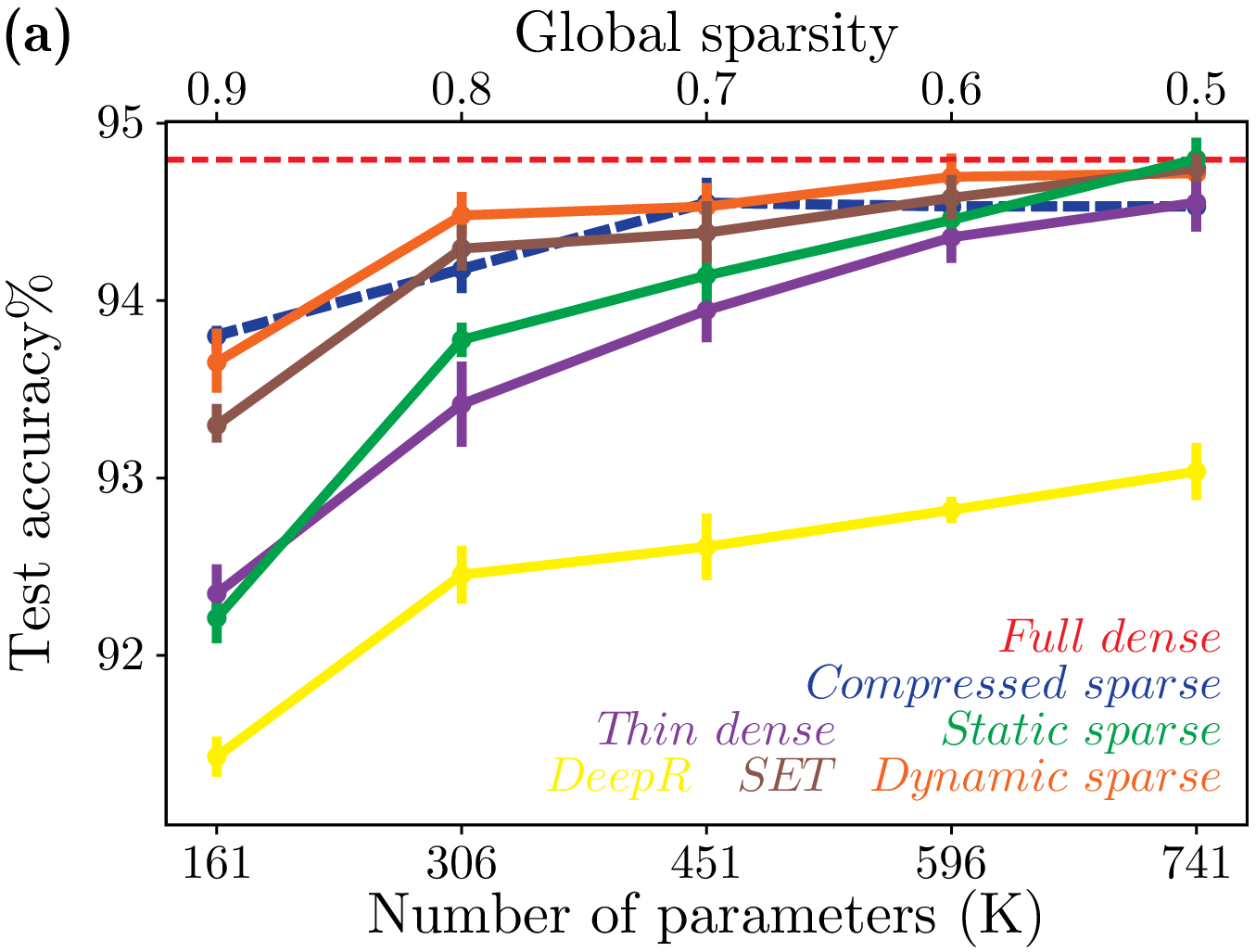} 
    \label{fig:cf10_accuracy}
  \end{subfigure}
  \quad
  \begin{subfigure}[t]{0.45\textwidth}
    \phantomsubcaption
    \includegraphics[width=\textwidth]{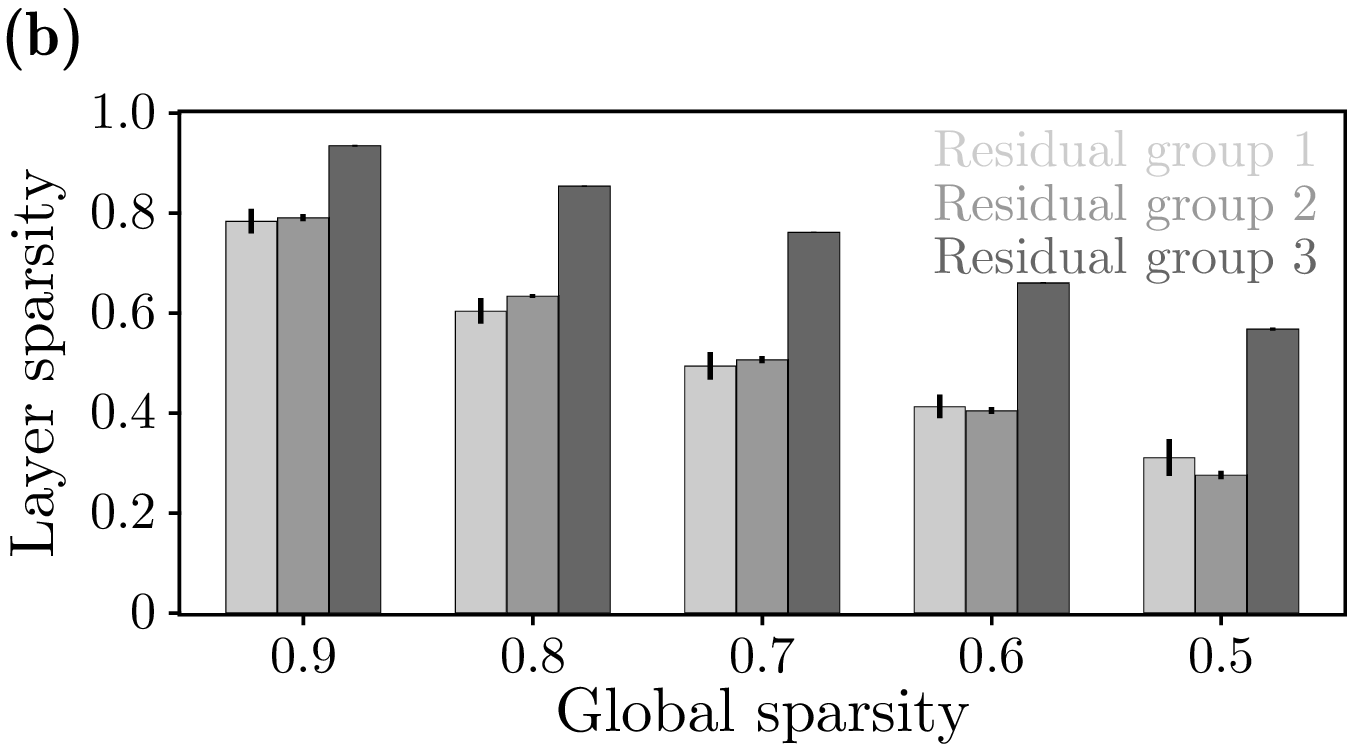} 
    \label{fig:cf10_block_sparsity}
  \end{subfigure}
  \caption{\footnotesize
    WRN-28-2 on CIFAR10.
    (\subref{fig:cf10_accuracy}) 
    Test accuracy plotted against number of trainable parameters in the sparse models for different methods. Dashed lines are used for the full dense model and for models obtained through compression, whereas all methods that maintain a constant parameter count throughout training and inference are represented by solid lines. Circular symbols mark the median of 5 runs, and error bars are the standard deviation. Parameter counts include all trainable parameters, i.e, parameters in sparse tensors plus all other dense tensors, such as those of batch normalization layers. 
    (\subref{fig:cf10_block_sparsity}) 
    Breakdown of the final sparsities of the parameter tensors in the three residual blocks that emerged from our dynamic sparse parameterization algorithm (Algorithm~\ref{algo:main}) at different levels of global sparsity. 
}
\end{figure}

\paragraph{WRN-28-2 on CIFAR10}
We ran experiments on a Wide Resnet model WRN-28-2~\citep{Zagoruyko2016} trained to classify CIFAR10 images (see Appendix~B for details of implementation). 
We varied global sparsity levels and evaluated test accuracy of different training methods based on dynamic and static reparameterization. 
As shown in Figure~\ref{fig:cf10_accuracy}, \emph{static sparse} and \emph{thin dense} significantly underperformed \emph{compressed sparse} model as expected, whereas our \emph{dynamic sparse} performed on par or slightly better than \emph{compressed sparse} on average. 
\emph{DeepR} significantly underperformed all other method. 
Though \emph{SET} was generally on par with \emph{compressed sparse} and \emph{dynamic sparse} at low sparsity levels, it underperformed \emph{dynamic sparse} at high sparsity levels. 
Even though the statically reparameterized models \emph{static sparse} and \emph{thin dense} were trained for twice the number of epochs, they still failed to reach the accuracy of \emph{dynamic sparse} or \emph{SET}. 
Note that \emph{thin dense} had even more trainable free parameters than all sparse models (see Section~\ref{sec:meth}).     

Further, we inspected the layer-wise sparsity patterns that emerged from automatic parameter reallocation across layers (Algorithm~\ref{algo:main}) during \emph{dynamic sparse} training.  
We observed consistent patterns at different sparsity levels: (a) larger parameter tensors tended to be sparser than smaller ones, and (b) deeper layers tended to be sparser than shallower ones. 
Figure~\ref{fig:cf10_block_sparsity} shows a breakdown of the final sparsity levels of different residual blocks at different sparsity levels.

\begin{figure}[h]
  \centering
  \begin{subfigure}[b]{0.45\textwidth}
    \phantomsubcaption
    \includegraphics[width=\textwidth]{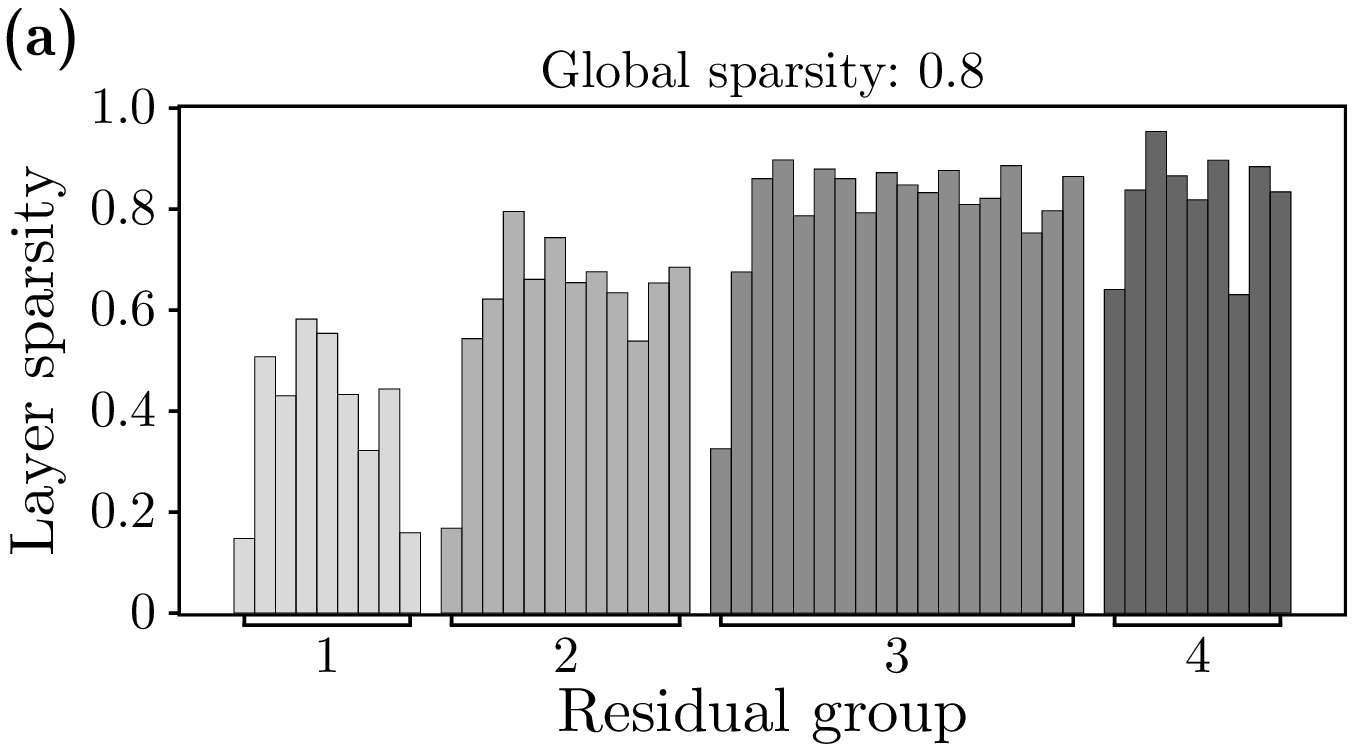} 
    \label{fig:resnet_layersparsity08}
  \end{subfigure}
  \quad
  \begin{subfigure}[b]{0.45\textwidth}
    \phantomsubcaption
    \includegraphics[width=\textwidth]{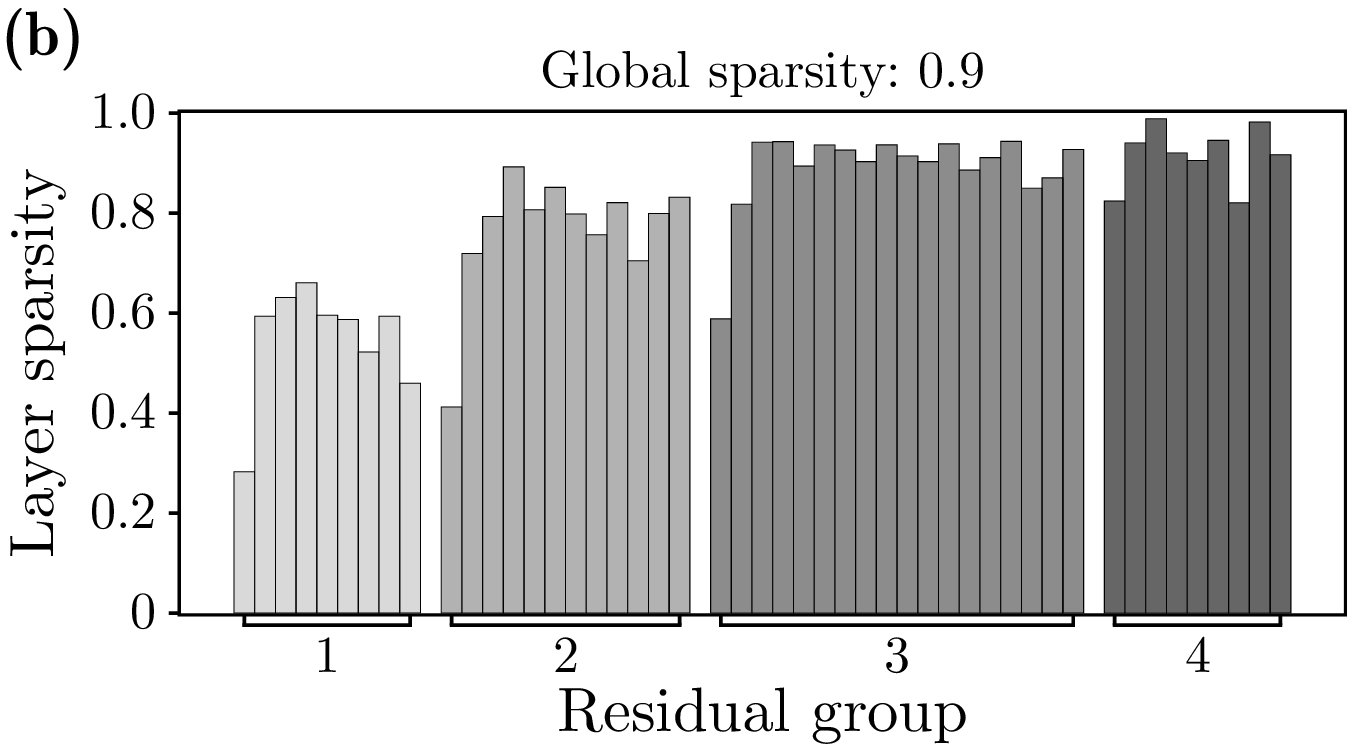} 
    \label{fig:resnet_layersparsity09}
  \end{subfigure}
  \caption{\footnotesize
    layer-wise breakdown of the final parameter tensor sparsities of Resnet-50 trained on Imagenet.
    (\subref{fig:resnet_layersparsity08}) 
    At overall sparsity $0.8$.
    (\subref{fig:resnet_layersparsity09}) 
    At overall sparsity of $0.9$. 
  }
  \label{fig:resnet_sparsity}
\end{figure}

\begin{table*}[t]
\caption{
  Test accuracy\% (top-1, top-5) of Resnet-50 trained on Imagenet 
}
\label{tb:imagenet_accuracy}
\vspace{-3mm}\centering
\setlength\tabcolsep{5pt}
\begin{tabular}{ l | l | l | cc | cc | cc }
  \toprule
  \multicolumn{3}{r|}{Final overall sparsity (\# Parameters)}  & \multicolumn{2}{c|}{$0.8$ (7.3M)}  & \multicolumn{2}{c|}{$0.9$ (5.1M)} & \multicolumn{2}{c}{$0.0$ (25.6M)}\\
  \midrule
  \multirow{9}{*}{Reparameterization}
  &\multirow{3}{*}{
    \begin{tabular}{@{}l@{}}Static\end{tabular}
  }
  &\emph{Thin dense}               
  & \begin{tabular}{@{}c@{}}          72.4 \\           {[-2.5]}       \end{tabular}      
  & \begin{tabular}{@{}c@{}}          90.9 \\           {[-1.5]}       \end{tabular}      
  & \begin{tabular}{@{}c@{}}          70.7 \\           {[-4.2]}       \end{tabular}      
  & \begin{tabular}{@{}c@{}}          89.9 \\           {[-2.5]}       \end{tabular}      
  &
  \multirow{12}{*}{ 
    \begin{tabular}{@{}c@{}}          74.9 \\           {[0.0]}        \end{tabular} 
  } & \multirow{12}{*}{ 
    \begin{tabular}{@{}c@{}}          92.4 \\           {[0.0]}        \end{tabular} 
  }
  \\ 
  & &\emph{Static sparse}            
  & \begin{tabular}{@{}c@{}}          71.6 \\           {[-3.3]}       \end{tabular}      
  & \begin{tabular}{@{}c@{}}          90.4 \\           {[-2.0]}       \end{tabular}      
  & \begin{tabular}{@{}c@{}}          67.8 \\           {[-7.1]}       \end{tabular}      
  & \begin{tabular}{@{}c@{}}          88.4 \\           {[-4.0]}       \end{tabular}      
  \\ 
  \cmidrule{2-7}
  &\multirow{5}{*}{
    \begin{tabular}{@{}l@{}}Dynamic\end{tabular}
  }
  & \begin{tabular}{@{}l@{}}\emph{DeepR}\\~{ \citep{Bellec2017}}\end{tabular}            
  & \begin{tabular}{@{}c@{}}          71.7 \\           {[-3.2]}       \end{tabular}      
  & \begin{tabular}{@{}c@{}}          90.6 \\           {[-1.8]}       \end{tabular}      
  & \begin{tabular}{@{}c@{}}          70.2 \\           {[-4.7]}       \end{tabular}      
  & \begin{tabular}{@{}c@{}}          90.0 \\           {[-2.4]}       \end{tabular}      
  \\ 
  & &\begin{tabular}{@{}l@{}}\emph{SET}\\~{ \citep{Mocanu2018}}\end{tabular}            
  & \begin{tabular}{@{}c@{}}          72.6 \\           {[-2.3]}       \end{tabular}      
  & \begin{tabular}{@{}c@{}}          91.2 \\           {[-1.2]}       \end{tabular}      
  & \begin{tabular}{@{}c@{}}          70.4 \\           {[-4.5]}       \end{tabular}      
  & \begin{tabular}{@{}c@{}}          90.1 \\           {[-2.3]}       \end{tabular}      
  \\ 
  & &\begin{tabular}{@{}l@{}}\emph{Dynamic sparse}\\~{ (Ours)}\end{tabular}    
  & \begin{tabular}{@{}c@{}}      \bf 73.3 \\           {[{\bf -1.6}]}       \end{tabular}      
  & \begin{tabular}{@{}c@{}}      \bf 92.4 \\           {[{\bf \ 0.0}]}        \end{tabular}      
  & \begin{tabular}{@{}c@{}}      \bf 71.6 \\           {[{\bf -3.3}]}       \end{tabular}      
  & \begin{tabular}{@{}c@{}}      \bf 90.5 \\           {[{\bf -1.9}]}       \end{tabular}      
  \\ 
  \cmidrule{1-7}
  \multicolumn{2}{l|}{\multirow{5}{*}{Compression}}
  &\begin{tabular}{@{}l@{}}\emph{Compressed sparse}\\~{ \citep{Zhu2017}}\end{tabular}        
  & \begin{tabular}{@{}c@{}}          73.2 \\           {[-1.7]}       \end{tabular}      
  & \begin{tabular}{@{}c@{}}          91.5 \\           {[-0.9]}       \end{tabular}      
  & \begin{tabular}{@{}c@{}}          70.3 \\           {[-4.6]}       \end{tabular}      
  & \begin{tabular}{@{}c@{}}          90.0 \\           {[-2.4]}       \end{tabular}      
  \\ 
  \cmidrule{3-9}\morecmidrules\cmidrule{3-9}
  \multicolumn{2}{l|}{} 
  & \begin{tabular}{@{}l@{}}\emph{ThiNet}\\~{ \citep{Luo2017}}\end{tabular}  
  & \begin{tabular}{@{}c@{}}          \textit{68.4} \\           {[\textit{-4.5}]}       \end{tabular}  
  & \multicolumn{1}{c}{
    \begin{tabular}{@{}c@{}}          \textit{88.3} \\           {[\textit{-2.8}]}       \end{tabular}
  } 
  & \multicolumn{4}{l}{(at 8.7M parameter count)}\\ 
  \multicolumn{2}{l|}{} 
  & \begin{tabular}{@{}l@{}}\emph{SSS}\\~{ \citep{Huang2017}}\end{tabular}  
  & \begin{tabular}{@{}c@{}}          \textit{71.8} \\           {[\textit{-4.3}]}       \end{tabular}  
  & \multicolumn{1}{c}{
    \begin{tabular}{@{}c@{}}          \textit{90.8} \\           {[\textit{-2.1}]}       \end{tabular}
  } 
  & \multicolumn{4}{l}{(at 15.6M parameter count)}\\ 
  \bottomrule
\end{tabular} 
\captionsetup{singlelinecheck=off,font=footnotesize,width=0.92\textwidth}
\caption*{
  Numbers in square brackets are differences from the \emph{full dense} baseline.  
  Romanized numbers are results of our experiments, and italicized ones taken directly from the original paper.
  Performance of two structured pruning methods, \emph{ThiNet} and \emph{Sparse Structure Selection} (\emph{SSS}), are also listed for comparison (below the double line, see Appendix~D for a discussion of their relevance); note the difference in parameter counts.
}\end{table*}

\paragraph{Resnet-50 on Imagenet}
Next, we trained the Resnet-50 bottleneck architecture~\citep{He2015} on Imagenet (see Appendix~B for details of implementation). 
We ran experiments at two global sparsity levels, $0.8$ and $0.9$ (Table~\ref{tb:imagenet_accuracy}). 
Models obtained by our (\emph{dynamic sparse}) method outperformed all dynamic and static reparameterization baseline methods, slightly outperforming \emph{compressed sparse} models obtained through post-training compression of a large dense model. 
In Table~\ref{tb:imagenet_accuracy}, we also list two additional representative methods of \emph{structured} pruning (see Appendix~D), \emph{ThiNet}~\citep{Luo2017} and \emph{Sparse Structure Selection}~\citep{Huang2017}, which, consistent with recent criticisms~\citep{Liu2018}, underperformed static dense baselines. 
Similar to \emph{dynamic sparse} WRN-28-2, reliable sparsity patterns across parameter tensors in different layers emerged from dynamic parameter reallocation during training, displaying the same empirical trends described above (Figure~\ref{fig:resnet_sparsity}).

\begin{table}[!b]
\caption{
  Wall-clock training time comparison 
}
\vspace{-3mm}\centering
\setlength\tabcolsep{5pt}
\begin{tabular}{ l | c | c}
  \toprule
  &     \begin{tabular}{@{}c@{}} WRN-28-2 on \\ CIFAR10\end{tabular}   & \begin{tabular}{@{}c@{}} Resnet-50 on \\ Imagenet\end{tabular} \\
  \midrule
  \emph{DeepR}           & 4.466 $\pm$ 0.358     & 5.636 $\pm$ 0.218 \\
  \emph{SET}             & 1.087 $\pm$ 0.049     & 1.009 $\pm$ 0.002 \\
  \emph{Dynamic sparse} (ours)  & \bf 1.083 $\pm$ 0.051     & \bf 1.005 $\pm$ 0.004 \\
  \bottomrule
\end{tabular} 
\captionsetup{singlelinecheck=off,font=footnotesize}
\caption*{
  Median\,$\pm$\,standard deviation of wall-clock training epoch times (from 25 epochs) for WRN-28-2 and Resnet-50 for different dynamic reparameterization methods. 
  Results are relative ratios to the epoch time of a sparse network trained without dynamic parameter reallocation. 
  WRN-28-2 is trained on a single, while Resnet-50 on four, Nvidia TITAN Xp GPU(s). 
}
\label{tb:comp_overhead}
\end{table}

\paragraph{Computational overhead of dynamic parameter reallocation}
We assessed the additional computational cost incurred by dynamic parameter reallocation steps (Algorithm~\ref{algo:main}) during training, and compared ours with existing dynamic sparse reparameterization techniques, \emph{DeepR} and \emph{SET} (Table~\ref{tb:comp_overhead}).  
Because both \emph{SET} and \emph{dynamic sparse} reallocate parameters only intermittently (every few hundred training iterations), the computational overhead was negligible for the experiments presented here\footnote{
  Because of the rather negligible overhead, the reduced operation count thanks to the elimination of sorting operations did not amount to a substantial improvement over SET on GPUs. 
  Our method's another advantage over SET lies in its ability to produce better sparse models and to reallocate free parameters automatically (see Appendix~F).
}.
\emph{DeepR}, however, requires adding noise to gradient updates as well as reallocating parameters every training iteration, leading to a significantly larger overhead.

\begin{figure}[h]
  \captionsetup{aboveskip=0pt,belowskip=0pt}
    \centering
    \begin{subfigure}[b]{0.45\textwidth}
      \includegraphics[width=\textwidth]{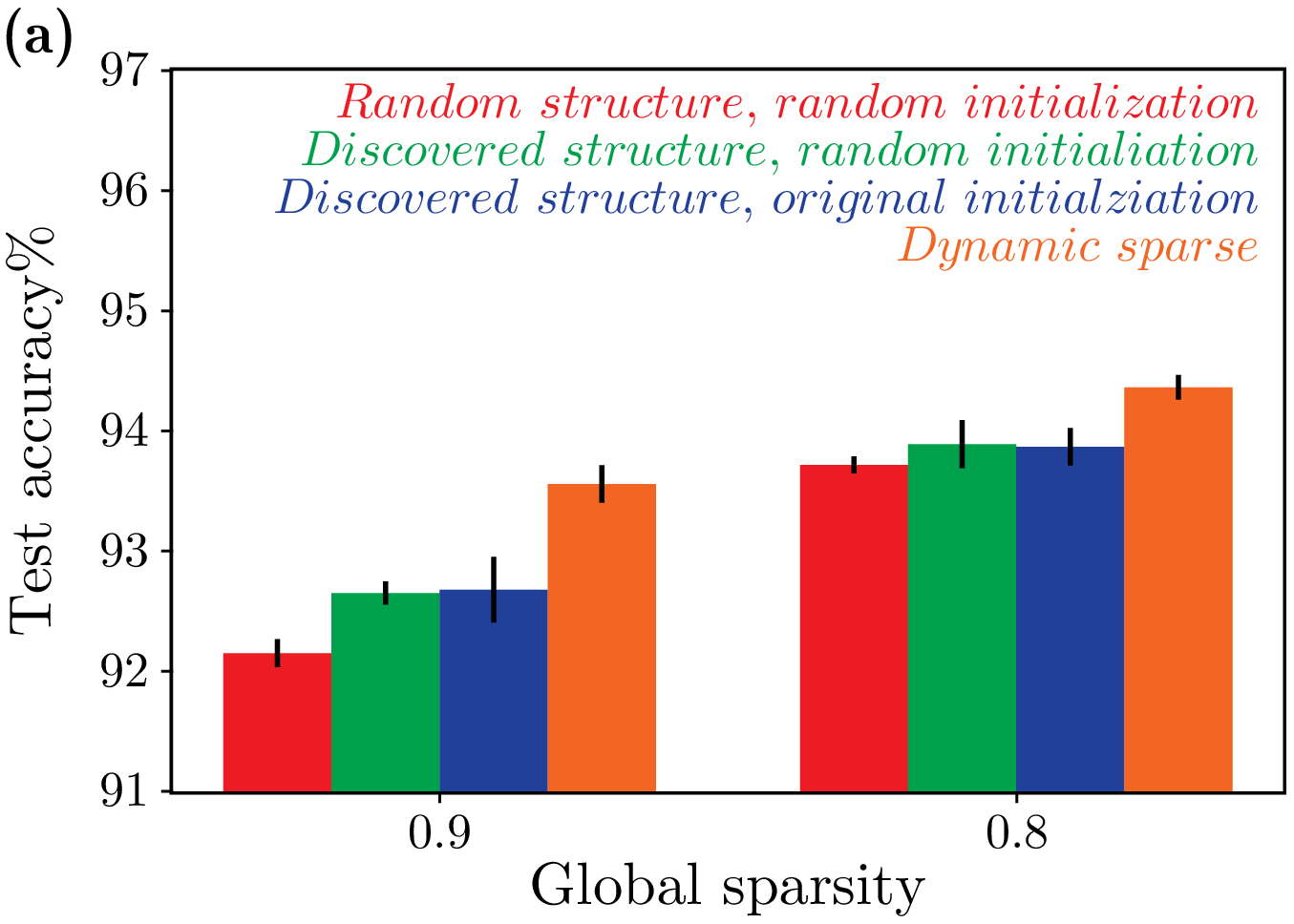} 
      \phantomsubcaption
      \label{fig:tickets_a}
    \end{subfigure}
    \quad
    \begin{subfigure}[b]{0.45\textwidth}
      \includegraphics[width=\textwidth]{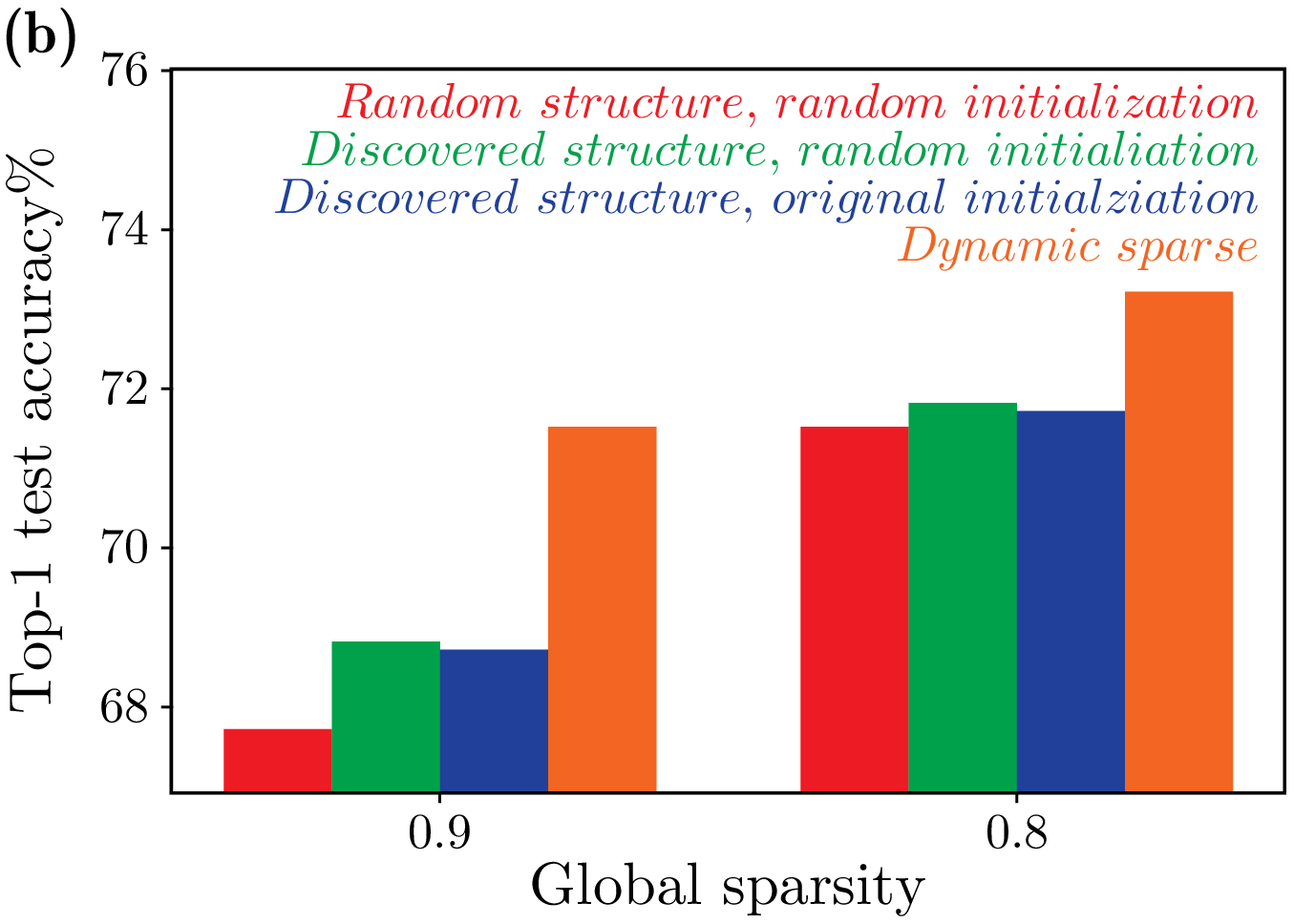} 
      \phantomsubcaption
      \label{fig:tickets_b}
    \end{subfigure}
    \caption{
       \footnotesize
       Comparison of training using our dynamic reparameterization method against training a number of related statically parameterized networks. All statically parameterized networks were trained for double the number of epochs used by our method. 
       (\subref{fig:tickets_a}) WRN-28-2 on CIFAR10. Mean and standard deviation from 5 runs.
       (\subref{fig:tickets_b}) Resnet-50 on Imagenet. Single run. 
     }
  \end{figure}
  
  \begin{figure}[h]
  \captionsetup{aboveskip=0pt,belowskip=0pt}
    \centering
      \includegraphics[width=0.45\textwidth]{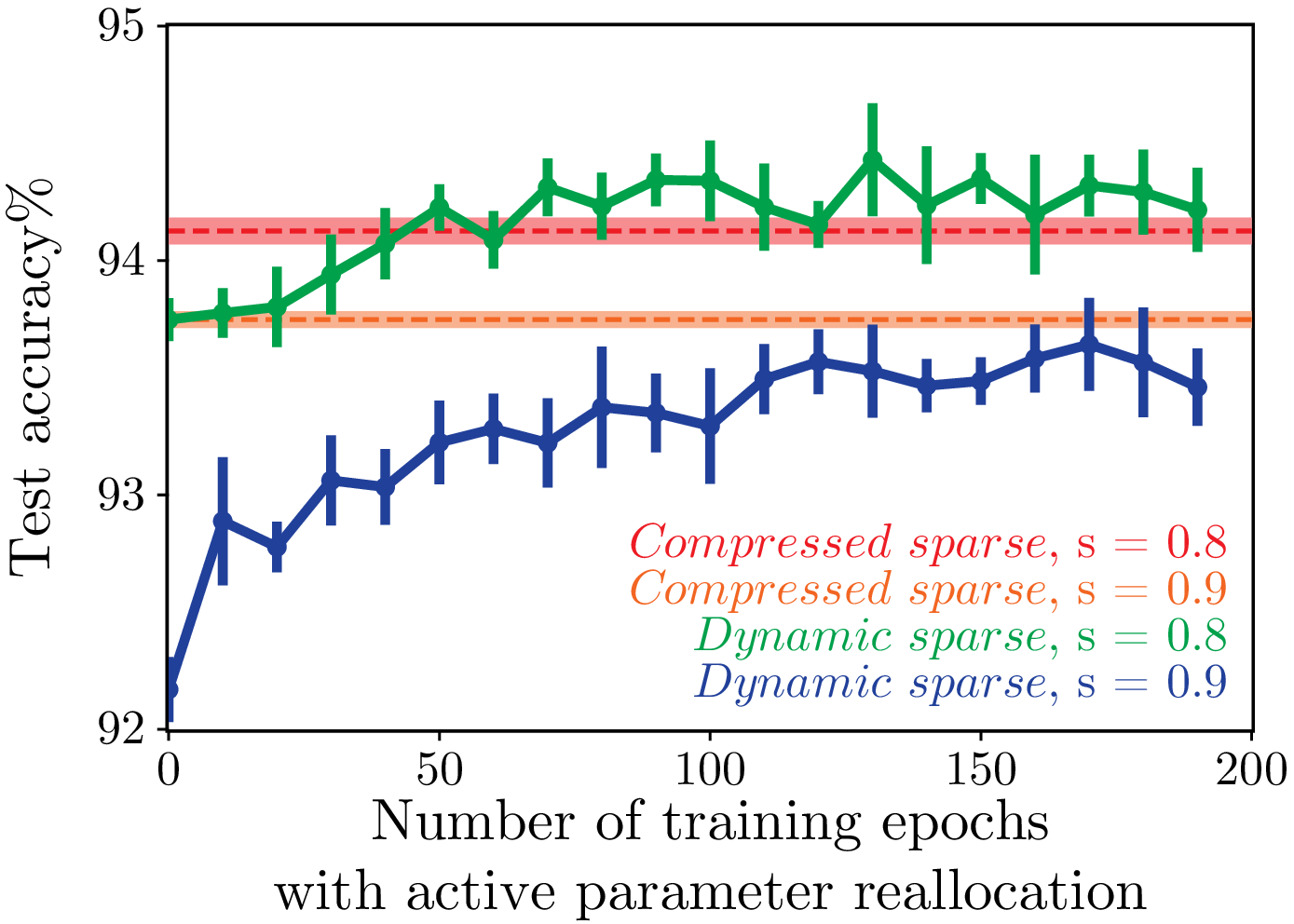} 
    \caption{\footnotesize
      Test accuracies of sparse WRN-28-2 trained on CIFAR10 when dynamic parameter reallocation was switched off at certain epochs. Results are shown for two global sparsity levels: $0.8$ and $0.9$. Horizontal bands indicate the accuracy of the \emph{compressed sparse} baselines where the band widths represent the standard deviation. For all data points, we ran training for 200 epochs, regardless of when dynamic parameter reallocation was stopped. Mean and standard deviation from 5 independent runs. 
    }
    \label{fig:early_stopping}
  \end{figure}

\paragraph{Understanding the effects of dynamic parameter reallocation}
Why did dynamic parameter reallocation yield sparse models that generalize better than static models trained \textit{de novo}?  
To address this question, we investigated whether our method discovered more trainable sparse network structures, following the reasoning of the recently proposed ``lottery ticket'' hypothesis~\citep{Frankle2018}. 

First, we did the following experiment with WRN-28-2 trained on CIFAR10: after training with \emph{dynamic sparse} method, we retained the final network sparseness pattern (i.e. positions of non-zero entries in all sparse parameter tensors), and then randomly re-initialized this network and re-trained with its structure fixed (Figure~\ref{fig:tickets_a}). 
It failed to reach comparable accuracy, suggesting that the sparse connectivity discovered by our method is not sufficient to explain the high generalization performance.

Next, we asked whether the particular weight initialization of the sparse network in addition to its sparseness structure led to high accuracies~\citep{Frankle2018}. 
We used the final network structure as described above, and re-initialized it with the exact same initial values used in the original training. 
As shown in Figure~\ref{fig:tickets_a}, the combination of final structure and original initialization still fell significantly short of the level of accuracy achieved by training with dynamic parameter reallocation, not significantly different from training the same network with random re-initialization. 

For control, we also show the performance of \emph{static sparse} models where the sparse network structure and its initialization were both random (Figure~\ref{fig:tickets_a}), which, not surprisingly, performed the worst. 
Similar trends were observed for Resnet-50 trained on Imagenet (Figure~\ref{fig:tickets_b}). 
All static networks, in all sparseness pattern and re-initialization configurations, were trained for double the number of epochs used for dynamic training. 

These results suggest that the dynamic evolution of the network through parameter reallocation is crucial to effective learning, because the superior generalization performance cannot be solely attributed to the network's structure, nor to its initialization, nor to a combination of the two. 

Finally, to investigate whether the convergence of sparse network structures and that of parameter values had similar time courses, we did extra experiments with WRN-28-2, where at various stages during training, we stopped dynamic parameter reallocation, fixing the network structure while continuing the optimization of parameter values. 
As shown in Figure~\ref{fig:early_stopping}, dynamic parameter reallocation does not need to be active for the entire course of training, but only for some initial epochs.
This suggests the network structure converges faster than the network parameters, which might be exploited in practice to further reduce computational cost.

%% file: discussion.tex
\section{Discussion}
\label{discussion}

In this work, we addressed the following problem: given a small, fixed budget of parameters for a deep residual CNN throughout training time, how to train it to yield the best generalization performance. 
We showed that training with dynamic parameter reallocation can achieve significantly better accuracies than static reparameterization at the same model size. 
Dynamic sparse reparameterization techniques have received relatively little attention to date, two existing methods (\emph{SET} and \emph{DeepR}) being applied only to relatively small and shallow networks. 
We proposed a dynamic parameterization method that adaptively reallocates free parameters across the network based on a simple heuristic during training. 
Our method yielded sparse models that generalize better than those produced by previous dynamic parameterization methods and outperformed all the static reparameterization methods we benchmarked against\footnote{
Additionally, we show that our method outperformed a static dense reparameterization method \emph{HashedNet}~\cite{Chen2015a} (Appendix~C), and that it is also able to train networks at extreme sparsity levels where previous static and dynamic parameterization methods often fail catastrophically (Appendix~F). 
}. 

High-performance sparse networks are often obtained by post-training pruning of dense networks. 
A number of recent studies have attempted direct training of sparse networks using \textit{post hoc} information obtained from a pruned model. 
\cite{Liu2018} argued that the sparse structure of the pruned model alone suffices to yield high accuracy, i.e. training a model of the same structure, starting with random weights, almost always reaches comparable levels of accuracy as the pruned model. 
In contrast, \cite{Frankle2018} argued that a post-compression sparse network structure alone is not sufficient, but necessary, to attain high accuracy, which, as the authors argue, requires both the pruned network connectivity \emph{and} its initial weights when it was trained as part of the dense model pre-compression.
In our experiments, we found that neither the \textit{post hoc} sparseness pattern, nor the combination of connectivity and initialization, managed to explain the high performance of sparse networks produced by our \emph{dynamic sparse} training method. 
Thus, the value of dynamic parameter reallocation goes beyond discovering trainable sparse network structure: the evolutionary process of structural exploration itself seems helpful for SGD to converge to better weights. 
Extra work is needed to explain the mechanisms underlying this phenomenon. 
One hypothesis is that the discontinuous jumps in parameter space when parameters are reallocated across layers helped training escape sharp minima that generalize badly~\cite{Keskar2016}.

Structural degrees of freedom are qualitatively different from the degrees of freedom introduced by overparameterization. 
The latter directly expands the dimensionality of the parameter space in which SGD directly optimizes, whereas structural degrees of freedom are realized and explored using non-differentiable heuristics that only interact indirectly with the dynamics of gradient-based optimization, e.g. regularization pulling weights towards zero causing connections to be pruned. 
Our results suggest that, for residual CNNs under a given descriptive complexity (i.e. memory storage) constraint, it is better (in the sense of producing models that generalize better) to allocate some memory to describe and explore structural degrees of freedom, than to allocate all memory to conventional weights.  
This makes a potentially compelling case for hardware acceleration of sparse computations for more parameter efficient training.

Beside storage (spatial complexity), computational efficiency (temporal complexity) is also of primary concern. 
Current mainstream computing hardware architectures such as CPUs and GPUs cannot efficiently handle unstructured sparsity patterns. 
To maintain structured network configurations, various pruning techniques prune a trained model at the level of entire feature maps or layers. 
Emerging evidence suggests that the resulting pruned networks perform no better than directly-trained thin networks~\citep{Liu2018}, calling into question the value of such coarse-grained pruning. 
We show in Appendix~E additional results applying \emph{dynamic sparse} training at an intermediate level of structured sparseness, i.e. pruning $3\times3$ kernel slices. 
Imposing this sparseness structure led to significantly worse generalization, producing sparse networks performing on par with statically parameterized \emph{thin dense} networks trained for double the number of epochs.

In summary, we show in this paper that it is possible to train deep sparse CNNs directly to reach generalization performances comparable to those achieved by iterative pruning and fine-tuning of pre-trained large dense models. 
The performance level achieved by our proposed method is significantly higher than that achieved by training dense models of the same size. 
Our method is the first to reallocate free parameters effectively and automatically within and across layers.
Furthermore, we show that dynamic exploration of structural degrees of freedom during training is crucial to effective learning.
Our work does not contradict the common wisdom that extra degrees of freedom are helpful for training deep networks to achieve better generalization, but it suggests that adding and dynamically exploring structural degrees of freedom is often a more effective and efficient alternative than simply increasing the parameter counts of the model.

%% file: supplementary.tex
\appendix
\appendixpage
\addappheadtotoc

\begin{appendices} 

\section{A full description of the dynamic parameter reallocation algorithm}
\label{app:alg}

Algorithm~1 in the main text informally describes our parameter reallocation scheme. 
In this appendix, we present a more rigorous description of the algorithm. 

Let all reparameterized weight tensors in the original network be denoted by $\{\tW_l\}$, where $l=1,\cdots,L$ indexes layers. 
Let $N_l$ be the number of parameters in $\tW_l$, and $N = \sum_l N_l$ the total parameter count.  

Sparse reparameterize $\tW_l = g\left( \vphi_l; \vpsi_l \right)$, where function $g$ places components of parameter $\vphi_l$ into positions in $\tW_l$ indexed by $\vpsi_l \in \mathbf{\Psi}_{M_l}\left(\{1,\cdots,N_l\}\right)$ \footnote{
    By $\mathbf{\Psi}_p(Q) \triangleq \left\{ \sigma(\Psi) : \Psi \in 2^Q, \abs{\Psi} = p, \sigma \in \mathrm{S}_p \right\}$ we denote the set of all cardinality $p$ ordered subsets of finite set $Q$.
}, s.t. $W_{l,\psi_{l,i}}=\phi_{l,i},\forall i$ indexing components.
Let $M_l < N_l$ be the dimensionality of $\vphi_l$ and $\vpsi_l$, i.e. the number of non-zero weights in $\tW_l$.  
Define $s_l = 1 - \frac {M_l} {N_l}$ as the \emph{sparsity} of $\tW_l$.
Global sparsity is then defined as $s = 1 - \frac M N$ where $M = \sum_l M_l$.  

During the whole course of training, we kept global sparsity constant, specified by hyperparameter $s \in (0, 1)$.  
Reparameterization was initialized by uniformly sampling positions in each weight tensor at the global sparsity $s$, i.e.  
    $\vpsi_l^{(0)} \thicksim \mathcal{U} \left[ 
        \mathbf{\Psi}_{M_l^{(0)}} \left( \left\{ 1,\cdots,N_l \right\} \right) 
    \right], \forall l$,
where $M_l^{(0)} = \round{(1-s)N_l}$.  
Associated parameters $\vphi_l^{(0)}$ were randomly initialized.  

Dynamic reparameterization was done periodically by repeating the following steps during training:
\begin{enum}
    \item Train the model (currently reparameterized by $\left\{\left(\vphi_l^{(t)}, \vpsi_l^{(t)}\right)\right\}$) for $P$ batch iterations;
    \item Reallocate free parameters within and across weight tensors following Algorithm~\ref{alg} to arrive at new reparameterization $\left\{\left(\vphi_l^{(t+1)}, \vpsi_l^{(t+1)}\right)\right\}$.
\end{enum}

\begin{algorithm*}[t]
\caption{Reallocate free parameters within and across weight tensors}
\label{alg}
\algnewcommand\algorithmicinput{\textbf{Input:}}
\algnewcommand\Input{\item[\algorithmicinput]}
\algnewcommand\algorithmicoutput{\textbf{Output:}}
\algnewcommand\Output{\item[\algorithmicoutput]}
\algrenewcommand\alglinenumber[1]{\ttfamily\footnotesize #1}
\algnewcommand\algorithmicneed{\textbf{Need:}}
\algnewcommand\Need{\item[\algorithmicneed]}
\begin{algorithmic}[1]
    \Input $\left\{\left(\vphi_l^{(t)}, \vpsi_l^{(t)}\right)\right\}$, $M^{(t)}$ , $H^{(t)}$ 
        \Comment{From step $t$}
    \Output $\left\{\left(\vphi_l^{(t+1)}, \vpsi_l^{(t+1)}\right)\right\}$, $M^{(t+1)}$, $H^{(t+1)}$
        \Comment{To step $t+1$}
    \Need $N_p, \delta$ \Comment{Target number of parameters to be pruned and its fractional tolerance} 

    \For {$l \in \{1,\cdots,L\}$}
            \Comment{For each reparameterized weight tensor}
        \State $\Pi_l^{(t)} \gets \left\{ i: \abs{\phi_{l,i}^{(t)}} < H^{(t)} \right\}$
            \Comment{Indices of subthreshold components of $\vphi_l^{(t)}$ to be pruned}
        \State $\left(K_l^{(t)}, R_l^{(t)}\right) \gets \left(\abs{\Pi_l^{(t)}}, M_l^{(t)}-\abs{\Pi_l^{(t)}}\right)$
            \Comment{Numbers of pruned and surviving weights}
    \EndFor
    \If {$\sum_l K_l^{(t)} < (1-\delta) N_p$} 
            \Comment{Too few parameters pruned}
        \State $H^{(t+1)} \gets 2 H^{(t)}$
            \Comment{Increase pruning threshold}
    \ElsIf {$\sum_l K_l^{(t)} > (1+\delta) N_p$}
            \Comment{Too many parameters pruned}
        \State $H^{(t+1)} \gets \frac{1}{2} H^{(t)}$
            \Comment{Decrease pruning threshold}
    \Else   \Comment{A proper number of parameters pruned}
        \State $H^{(t+1)} \gets H^{(t)}$
            \Comment{Maintain pruning threshold}
    \EndIf
    \For {$l \in \{1,\cdots,L\}$}
            \Comment{For each reparameterized weight tensor}
        \State $G_l^{(t)} \gets \Round{\frac {R_l^{(t)}} {\sum_l R_l^{(t)}} \sum_l K_l^{(t)}}$
            \Comment{Redistribute parameters for growth}
        \State $\tilde\vpsi_l^{(t)} \thicksim 
            \mathcal{U} \left[
                \mathbf{\Psi}_{G_l^{(t)}} \left(
                    \left\{ 1,\cdots,N_l \right\} 
                    \setminus 
                    \left\{ \psi_{l,i}^{(t)} \right\}
                \right)
            \right]$
            \Comment{Sample zero positions to grow new weights}
        \State $M_l^{(t+1)} \gets M_l^{(t)} - K_l^{(t)} + G_l^{(t)}$
            \Comment{New parameter count}
        \State $\left(\vphi_l^{(t+1)}, \vpsi_l^{(t+1)}\right) \gets
            \left( \left[\vphi_{l, i \notin \Pi_l^{(t)}}^{(t)}, \vzero\right], \left[\vpsi_{l, i \notin \Pi_l^{(t)}}^{(t)}, \tilde\vpsi_l^{(t)}\right] \right)$
            \Comment{New reparameterization}
    \EndFor
\end{algorithmic}
\end{algorithm*}

The adaptive reallocation is in essence a two-step procedure: a global pruning followed by a tensor-wise growth.
Specifically our algorithm has the following key features:
\begin{ls}
    \item Pruning was based on magnitude of weights, by comparing all parameters to a global threshold $H$, making the algorithm much more scalable than methods relying on layer-specific pruning. 
    \item We made $H$ adaptive, subject to a simple setpoint control dynamics that ensured roughly $N_p$ weights to be pruned globally per iteration. 
    This is computationally cheaper than pruning exactly $N_p$ smallest weights, which requires sorting all weights in the network.  
    \item Growth was by uniformly sampling zero weights and tensor-specific, thereby achieving a reallocation of parameters across layers.  
    The heuristic guiding growth is
    \begin{align}\label{eq:growth_heuristic}
        G_l^{(t)} = \ROUND{\frac {R_l^{(t)}} {\sum_l R_l^{(t)}} \sum_l K_l^{(t)}} ,
    \end{align}
    where $K_l^{(t)}$ and $R_l^{(t)} = M_l^{(t)} - K_l^{(t)}$ are the pruned and surviving parameter counts, respectively. 
    This rule allocated more free parameters to weight tensors with more surviving entries, while keeping the global sparsity the same by balancing numbers of parameters pruned and grown~\footnote{
        Note that an exact match is not guanranteed due to rounding errors in Eq.~\ref{eq:growth_heuristic} and the possibility that $M_l^{(t)}-K_l^{(t)}+G_l^{(t)}>N_l$, i.e. free parameters in a weight tensor exceeding its dense size after reallocation. 
        We added an extra step to redistribute parameters randomly to other tensors in these cases, thereby assuring an exact global sparsity.
    }.
\end{ls}

The entire procedure can be fully specified by hyperparameters $\left(s, P, N_p, \delta, H^{(0)}\right)$.

\section{Details of implementation}
\label{app:details}

We implemented all models and reparameterization mechanisms using \texttt{pytorch}. 
Experiments were run on GPUs, and all sparse tensors were represented as dense tensors filtered by a binary mask~\footnote{
  This is a mere implementational choice for ease of experimentation given available hardware and software, which did not save memory because of sparsity.  
  With computing substrate optimized for sparse linear algebra, our method is duly expected to realize the promised memory efficiency.  
}.  
Source code to reproduce all experiments is available in the anonymous repository: \url{https://github.com/IntelAI/dynamic-reparameterization}.

\begin{table*}[!t]
\caption{Hyperparameters for all experiments presented in the paper}
\label{tb:hyperparam}
\vspace{-3mm}\centering
\centering
\setlength\tabcolsep{5pt}
\begin{tabular}{l | r r | r r | r r }
  \toprule
    Experiment 
    & \multicolumn{2}{c|}{ 
      \begin{tabular}[t]{@{}c}LeNet-300-100 \\ on MNIST\end{tabular} 
    } 
    & \multicolumn{2}{c|}{ 
      \begin{tabular}[t]{@{}c}WRN-28-2 \\ on CIFAR10\end{tabular} 
    }
    & \multicolumn{2}{c}{ 
      \begin{tabular}[t]{@{}c}Resnet-50 \\ on Imagenet\end{tabular} 
    } \\ \midrule \midrule
  
  \multicolumn{7}{c} {Hyperparameters for training} \\ \midrule 
  Number of training epochs
    & \multicolumn{2}{r|}{100}  
    & \multicolumn{2}{r|}{200}  
    & \multicolumn{2}{r}{100} 
    \\ \midrule
  Mini-batch size       
    & \multicolumn{2}{r|}{100}  
    & \multicolumn{2}{r|}{100}  
    & \multicolumn{2}{r}{256} 
    \\ \midrule  
  \begin{tabular}[t]{@{}l}Learning rate schedule \\ (epoch range: learning rate)\end{tabular} 
    & \begin{tabular}[t]{@{}r@{}@{}@{}}
        1 - 25:\\ 
        26 - 50: \\ 
        51 - 75: \\
        76 - 100: \\
      \end{tabular} 
    & \begin{tabular}[t]{@{}r@{}@{}@{}}
        0.100 \\ 
        0.020 \\ 
        0.040 \\
        0.008 \\
      \end{tabular}  
    & \begin{tabular}[t]{@{}r@{}@{}@{}}
        1 - 60:\\ 
        61 - 120: \\ 
        121 - 160: \\
        161 - 200: \\
      \end{tabular}  
    & \begin{tabular}[t]{@{}r@{}@{}@{}}
        0.100 \\ 
        0.020 \\ 
        0.040 \\
        0.008 \\
      \end{tabular}  
    & \begin{tabular}[t]{@{}r@{}@{}@{}}
        1 - 30:\\ 
        31 - 60: \\ 
        61 - 90: \\
        91 - 100: \\
      \end{tabular}
    & \begin{tabular}[t]{@{}r@{}@{}@{}}
        0.1000 \\ 
        0.0100 \\ 
        0.0010 \\
        0.0001 \\
      \end{tabular}  
    \\  \midrule
  Momentum (Nesterov)
    & \multicolumn{2}{r|}{0.9}  
    & \multicolumn{2}{r|}{0.9}  
    & \multicolumn{2}{r}{0.9} 
    \\ \midrule  
  $L^1$ regularization multiplier 
    & \multicolumn{2}{r|}{0.0001}  
    & \multicolumn{2}{r|}{0.0}  
    & \multicolumn{2}{r}{0.0} 
    \\ \midrule
  $L^2$ regularization multiplier 
    & \multicolumn{2}{r|}{0.0}  
    & \multicolumn{2}{r|}{0.0005}  
    & \multicolumn{2}{r}{0.0001} 
    \\ \midrule \midrule

  \multicolumn{7}{c} {Hyperparameters for sparse compression (\emph{compressed sparse})~\citep{Zhu2017}} \\ \midrule 
  Number of pruning iterations ($T$) 
    & \multicolumn{2}{r|}{10}  
    & \multicolumn{2}{r|}{20}  
    & \multicolumn{2}{r}{20} 
    \\ \midrule  
  \begin{tabular}[t]{@{}l}Number of training epochs \\ between pruning iterations \end{tabular} 
    & \multicolumn{2}{r|}{2}  
    & \multicolumn{2}{r|}{2}  
    & \multicolumn{2}{r}{2} 
    \\ \midrule  
  Number of training epochs post-pruning
    & \multicolumn{2}{r|}{20}  
    & \multicolumn{2}{r|}{10}  
    & \multicolumn{2}{r}{10} 
    \\ \midrule  
  Total number of pruning epochs
    & \multicolumn{2}{r|}{40}  
    & \multicolumn{2}{r|}{50}  
    & \multicolumn{2}{r}{50} 
    \\ \midrule  
  \begin{tabular}[t]{@{}l}Learning rate schedule during pruning\\ (epoch range: learning rate)\end{tabular} 
    & \begin{tabular}[t]{@{}r@{}@{}}
        1 - 20:\\ 
        21 - 30: \\ 
        31 - 40: \\
      \end{tabular} 
    & \begin{tabular}[t]{@{}r@{}@{}}
        0.0200 \\ 
        0.0040 \\ 
        0.0008 \\
      \end{tabular}  
    & \begin{tabular}[t]{@{}r@{}@{}}
        1 - 25:\\ 
        25 - 35: \\ 
        36 - 50: \\
      \end{tabular}  
    & \begin{tabular}[t]{@{}r@{}@{}}
        0.0200 \\ 
        0.0040 \\ 
        0.0008 \\
      \end{tabular}  
    & \begin{tabular}[t]{@{}r@{}@{}}
        1 - 25:\\ 
        26 - 35: \\ 
        36 - 50: \\
      \end{tabular}
    & \begin{tabular}[t]{@{}r@{}@{}}
        0.0100 \\ 
        0.0010 \\ 
        0.0001 \\
      \end{tabular} \\  
  \midrule \midrule

  \multicolumn{7}{c} {Hyperparameters for dynamic sparse reparameterization (\emph{dynamic sparse}) (ours)} \\ \midrule
  Number of parameters to prune ($N_p$)  
    & \multicolumn{2}{r|}{600}  
    & \multicolumn{2}{r|}{20,000}  
    & \multicolumn{2}{r}{200,000} 
    \\ \midrule  
  Fractional tolerance of $N_p$ ($\delta$)  
    & \multicolumn{2}{r|}{0.1}  
    & \multicolumn{2}{r|}{0.1}  
    & \multicolumn{2}{r}{0.1} 
    \\ \midrule  
  Initial pruning threshold ($H^{(0)}$)   
    & \multicolumn{2}{r|}{0.001}  
    & \multicolumn{2}{r|}{0.001}  
    & \multicolumn{2}{r}{0.001} 
    \\ \midrule  
  \begin{tabular}[t]{@{}l}Reparameterization period ($P$) schedule \\ (epoch range: $P$) \end{tabular} 
    & \begin{tabular}[t]{@{}r@{}@{}@{}}
        1 - 25:\\ 
        26 - 50: \\ 
        51 - 75: \\
        76 - 100: \\
      \end{tabular} 
    & \begin{tabular}[t]{@{}r@{}@{}@{}}
        100 \\ 
        200 \\ 
        400 \\
        800 \\
      \end{tabular}  
    & \begin{tabular}[t]{@{}r@{}@{}@{}}
        1 - 25:\\ 
        26 - 80: \\ 
        81 - 140: \\
        141 - 200: \\
      \end{tabular}  
    & \begin{tabular}[t]{@{}r@{}@{}@{}}
        100 \\ 
        200 \\ 
        400 \\
        800 \\
      \end{tabular}  
    & \begin{tabular}[t]{@{}r@{}@{}@{}}
        1 - 25:\\ 
        26 - 50: \\ 
        51 - 75: \\
        76 - 100: \\
      \end{tabular}
    & \begin{tabular}[t]{@{}r@{}@{}@{}}
        1000 \\ 
        2000 \\ 
        4000 \\
        8000 \\
      \end{tabular}  
  \\  \midrule \midrule
  
  \multicolumn{7}{c} {Hyperparameters for Sparse Evolutionary Training (\emph{SET})~\citep{Mocanu2018}} \\ \midrule
  \begin{tabular}{@{}c@{}}Number of parameters to prune \\ at each re-parameterization step\end{tabular}
    & \multicolumn{2}{r|}{600}  
    & \multicolumn{2}{r|}{20,000}  
    & \multicolumn{2}{r}{200,000} 
    \\ \midrule  
  \begin{tabular}[t]{@{}l}Reparameterization period ($P$) schedule \\ (epoch range: $P$) \end{tabular} 
  & \begin{tabular}[t]{@{}r@{}@{}@{}}
        1 - 25:\\ 
        26 - 50: \\ 
        51 - 75: \\
        76 - 100: \\
      \end{tabular}
    & \begin{tabular}[t]{@{}r@{}@{}@{}}
        100 \\ 
        200 \\ 
        400 \\
        800 \\
      \end{tabular}  
    & \begin{tabular}[t]{@{}r@{}@{}@{}}
        1 - 25:\\ 
        26 - 80: \\ 
        81 - 140: \\
        141 - 200: \\
      \end{tabular}  
    & \begin{tabular}[t]{@{}r@{}@{}@{}}
        100 \\ 
        200 \\ 
        400 \\
        800 \\
      \end{tabular}  
    & \begin{tabular}[t]{@{}r@{}@{}@{}}
        1 - 25:\\ 
        26 - 50: \\ 
        51 - 75: \\
        76 - 100: \\
      \end{tabular}
    & \begin{tabular}[t]{@{}r@{}@{}@{}}
        1000 \\ 
        2000 \\ 
        4000 \\
        8000 \\
      \end{tabular}  
  \\  \midrule \midrule
  
  \multicolumn{7}{c} {Hyperparameters for Deep Rewiring (\emph{DeepR})~\citep{Bellec2017}} \\ \midrule
  $L^1$ regularization multiplier ($\alpha$)  
    & \multicolumn{2}{r|}{$10^{-4}$}  
    & \multicolumn{2}{r|}{$10^{-5}$}  
    & \multicolumn{2}{r}{$10^{-5}$} 
  \\ \midrule
  \begin{tabular}[t]{@{}l}Temperature ($T$) schedule \\ (epoch range: $T$) \end{tabular} 
    &  \begin{tabular}[t]{@{}r@{}@{}@{}}
        1 - 25:\\ 
        26 - 50: \\ 
        51 - 75: \\
        76 - 100: \\
      \end{tabular}
    &  \begin{tabular}[t]{@{}r@{}@{}@{}}
        $10^{-3}$ \\ 
        $10^{-4}$ \\ 
        $10^{-5}$ \\
        $10^{-6}$ \\
      \end{tabular}   
    & \begin{tabular}[t]{@{}r@{}@{}@{}}
        1 - 25:\\ 
        26 - 80: \\ 
        81 - 140: \\
        141 - 200: \\
      \end{tabular}  
    & \begin{tabular}[t]{@{}r@{}@{}@{}}
        $10^{-5}$ \\ 
        $10^{-8}$ \\ 
        $10^{-12}$ \\
        $10^{-15}$ \\
      \end{tabular}  
    & \begin{tabular}[t]{@{}r@{}@{}@{}}
        1 - 25:\\ 
        26 - 50: \\ 
        51 - 75: \\
        76 - 100: \\
      \end{tabular}
    & \begin{tabular}[t]{@{}r@{}@{}@{}}
        $10^{-5}$ \\ 
        $10^{-8}$ \\ 
        $10^{-12}$ \\
        $10^{-15}$ \\
      \end{tabular}  
    
  \\ \bottomrule

\end{tabular}  
\end{table*}

\paragraph{Training}
Hyperparameter settings for training are listed in the first block of Table~\ref{tb:hyperparam}.  
Standard mild data augmentation was used in all experiments for CIFAR10 (random translation, cropping and horizontal flipping) and for Imagenet (random cropping and horizontal flipping). The last linear layer of WRN-28-2 was always kept dense as it has a negligible number of parameters. The number of training epochs for the \emph{thin dense} and \emph{static sparse} baselines are double the number of training epochs shown in Table~\ref{tb:hyperparam}.

\paragraph{Sparse compression baseline}
We compared our method against iterative pruning methods~\citep{Han2015,Zhu2017}. We start from a full dense model trained with hyperparameters provided in the first block of Table~\ref{tb:hyperparam} and then gradually prune the network to a target sparsity in $T$ steps.
As in \cite{Zhu2017}, the pruning schedule we used was
\begin{align}
    s^{(t)} = s + (1-s)\left(1 - \frac{t}{T}\right)^3 ,
\end{align}
where $t = 0,1,\cdots,T$ indexes pruning steps, and $s$ the target sparsity reached at the end of training.  
Thus, this baseline (labeled as \emph{compressed sparse} in the paper) was effectively trained for more iterations (original training phase plus compression phase) than our \emph{dynamic sparse} method.
Hyperparameter settings for sparse compression are listed in the second block of Table~\ref{tb:hyperparam}.  

\paragraph{Dynamic reparameterization (ours)}
Hyperparameter settings for dynamic sparse reparameterization (Algorithm 1) are listed in the third block of Table~\ref{tb:hyperparam}.  

\paragraph{Sparse Evolutionary Training (SET)}
Because the larger-scale experiments here (WRN-28-2 on CIFAR10 and Resnet-50 on Imagenet) were not attempted by \cite{Mocanu2018}, no specific settings for reparameterization in these cases were available in the original paper.  
In order to make a fair comparison, we used the same hyperparameters as those used in our dynamic reparameterization scheme (third block in Table~\ref{tb:hyperparam}). 
At each reparameterization step, the weights in each layer were sorted by magnitude and the smallest fraction was pruned. 
An equal number of parameters were then randomly allocated in the same layer and initialized to zero. 
For control, the total number of reallocated weights at each step was chosen to be the same as our dynamic reparameterization method, as was the schedule for reparameterization.  

\paragraph{Deep Rewiring (DeepR)}
The fourth block in Table~\ref{tb:hyperparam} contain hyperparameters for the DeepR experiments. 
We refer the reader to~\cite{Bellec2017} for details of the deep rewiring algorithm and for explanation of the hyperparameters. We chose the DeepR hyperparameters for the different networks based on a parameter sweep.

\section{Comparison to dense reparameterization method \emph{HashedNet}}
\label{app:comp_tied}

We also compared our dynamic sparse reparameterization method to a number of static dense reparameterization techniques, e.g.~\cite{Denil2013,Yang2014,Moczulski2015,Sindhwani2015,Chen2015a,Treister2018}.  
Instead of sparsification, these methods impose structure on large parameter tensors by parameter sharing.  
Most of these methods have not been used for convolutional layers except for recent ones~\citep{Chen2015a,Treister2018}.  
We found that \emph{HashedNet}~\citep{Chen2015a} had the best performance over other static dense reparameterization methods, and also benchmarked our method against it.  
Instead of reparameterizing a parameter tensor with $N$ entries to a sparse one with $M<N$ non-zero components, \emph{HashedNet}'s reparameterization is to put $M$ free parameters into $N$ positions in the parameter through a random mapping from $\left\{1,\cdots,N\right\}$ to $\left\{1,\cdots,M\right\}$ computed by cheap hashing, resulting in a dense parameter tensor with shared components.  

\begin{figure}[t]
  \centering
  \begin{subfigure}[b]{0.45\textwidth}
    \includegraphics[width=\textwidth]{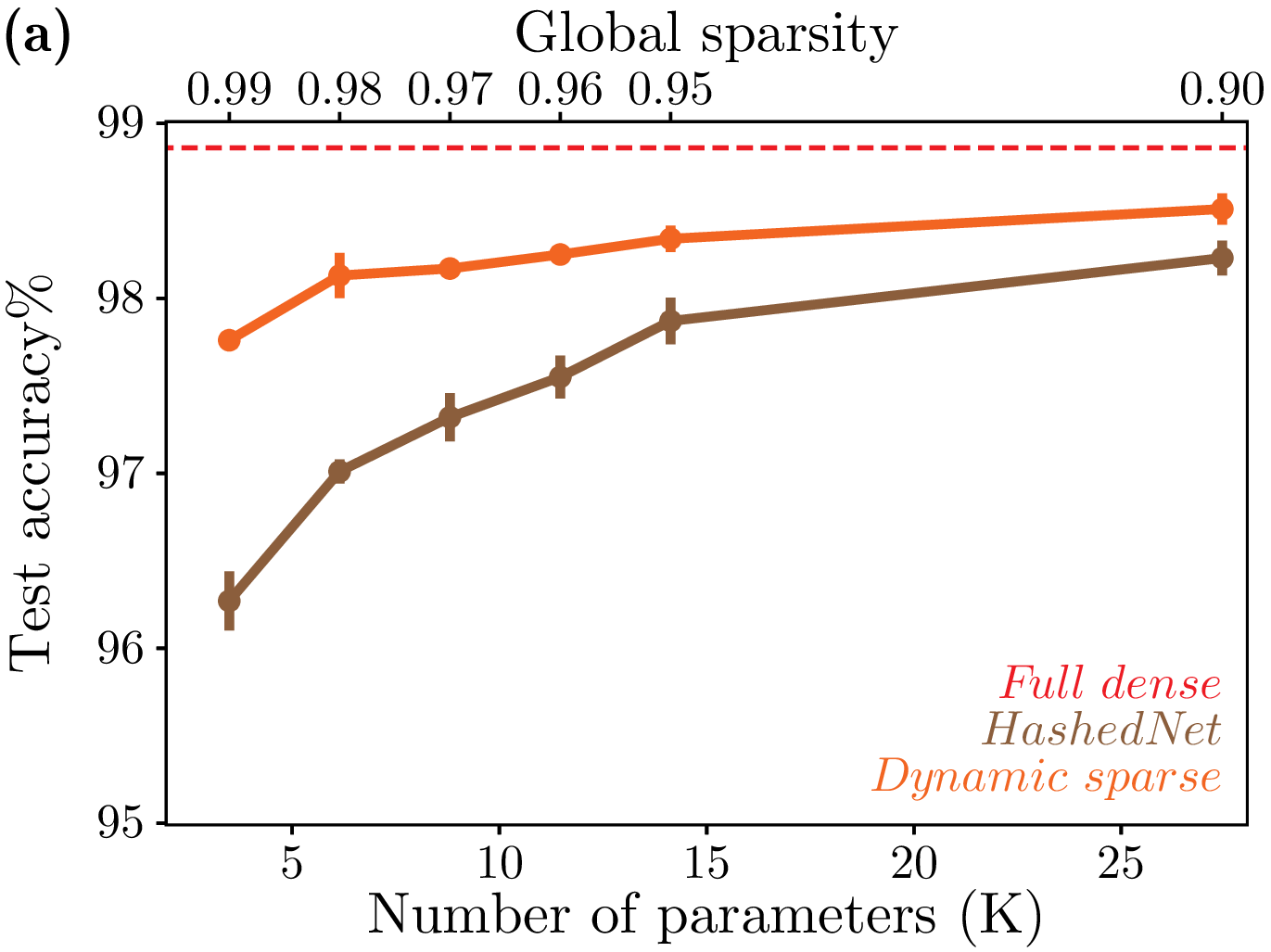} 
    \phantomsubcaption
    \label{fig:mnist_with_tied}
  \end{subfigure}
  \quad
  \begin{subfigure}[b]{0.45\textwidth}
    \includegraphics[width=\textwidth]{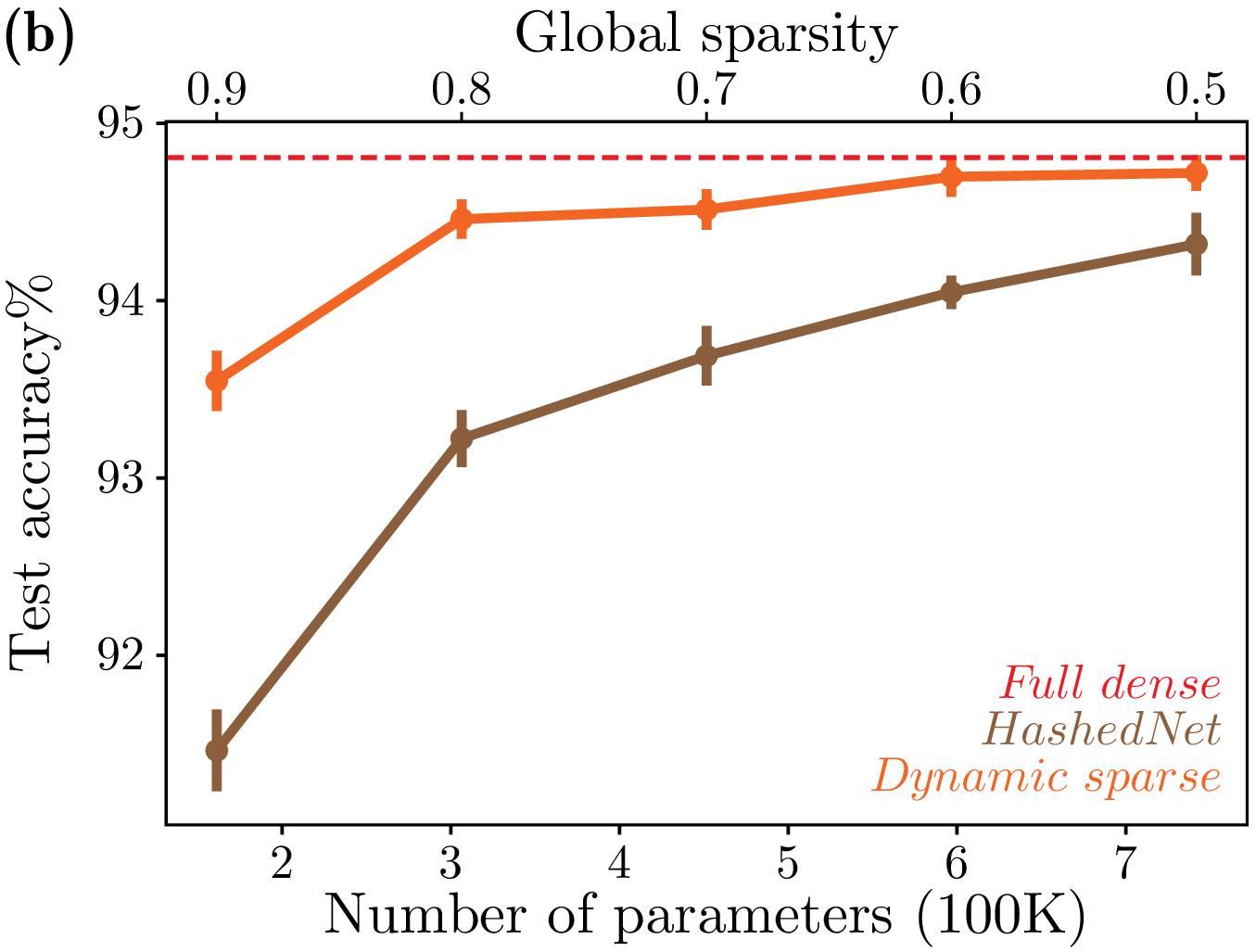} 
    \phantomsubcaption
    \label{fig:cifar_with_tied}
  \end{subfigure}
  \caption{\footnotesize
    Comparison to \emph{HashedNet}.
    (\subref{fig:mnist_with_tied}) 
    Test accuracy for LeNet-300-100-10 trained on MNIST.
    (\subref{fig:cifar_with_tied}) 
    Test accuracy for WRN-28-2 trained on CIFAR10. 
    Conventions same as in Figure~\ref{fig:mnist_accuracy}. 
  }
  \label{fig:tied_models}
\end{figure}

Results of LeNet-300-100-10 on MNIST are presented in Figure~\ref{fig:mnist_with_tied}, those of WRN-28-2 on CIFAR10 in Figure~\ref{fig:cifar_with_tied}, and those of Resnet-50 on Imagenet in Table~\ref{tb:imagenet_tied}.  
For a certain global sparsity $s$ of our method, we compare it against a \emph{HashedNet} with all reparameterized tensors hashed such that each had a fraction $1-s$ of unique parameters.  
We found that our method \emph{dynamic sparse} significantly outperformed \emph{HashedNet}.  

\begin{table*}[h]
\caption{
  Test accuracy\% (top-1, top-5) of Resnet-50 on Imagenet for \emph{dynamic sparse} vs. \emph{HashedNet}. Numbers in square brackets are differences from the \emph{full dense} baseline.  
}
\label{tb:imagenet_tied}
\vspace{-3mm}\centering
\setlength\tabcolsep{5pt}
\begin{tabular}{ l | ll | ll }
  \toprule
  Final global  sparsity (\# Parameters)
   & \multicolumn{2}{c|}{$0.8$ (7.3M)}  & \multicolumn{2}{c}{$0.9$ (5.1M)} \\
  \midrule
  \emph{HashedNet}                
  & 70.0 {[-4.9]}     
  & 89.6 {[-2.8]}     
  & 66.9 {[-8.0]}      
  & 87.4 {[-5.0]}      \\ 
  \emph{Dynamic sparse} (ours)    
  & {\bf 73.3} {[{\bf -1.6}]} 
  & {\bf 92.4} {[{\bf \ 0.0}]} 
  & {\bf 71.6} {[{\bf -3.3}]} 
  & {\bf 90.5} {[{\bf -1.9}]} \\ 
  \bottomrule
\end{tabular} 
  
\end{table*}

\section{A taxonomy of training methods that yield ``sparse'' deep CNNs}
\label{app:tax}

As an extension to Section 2 of the main text, here we elaborate on existing methods related to ours, how they compare with and contrast to each other, and what features, apart from effectiveness, distinguished our approach from all previous ones.  
We confine the scope of comparison to training methods that produce smaller versions (i.e. ones with fewer parameters) of a given modern (i.e. post-AlexNet) deep convolutional neural network model.  
We list representative methods in Table~\ref{tb:tax}. We classify these methods by three key features.

\begin{table*}[h]
\caption{
  Representative examples of training methods that yield ``sparse'' deep CNNs
}
\label{tb:tax}
\vspace{-3mm}\centering
\setlength\tabcolsep{5pt}
\begin{tabular}{ l | ccc }
  \toprule
  \bf Method 
  & \begin{tabular}{@{}c@{}}\bf Strict parameter budget \\ \bf throughout \\ \bf training and inference\end{tabular} 
  & \begin{tabular}{@{}c@{}}\bf Granularity \\ \bf of sparsity\end{tabular} 
  & \begin{tabular}{@{}c@{}}\bf Automatic \\ \bf layer sparsity\end{tabular} \\
   
  \midrule
  \begin{tabular}{@{}l@{}}Dynamic Sparse Reparameterization\\\quad(Ours)\end{tabular}             & yes     & non-structured    & yes   \\ 
  \begin{tabular}{@{}l@{}}Sparse Evolutionary Training (SET)\\\quad\citep{Mocanu2018}\end{tabular}& yes     & non-structured    & no  \\ 
  \begin{tabular}{@{}l@{}}Deep Rewiring (DeepR)\\\quad\citep{Bellec2017}\end{tabular}             & yes     & non-structured    & no  \\ 
  
  \midrule
  \begin{tabular}{@{}l@{}}NN Synthesis Tool (NeST)\\\quad\citep{Dai2017,Dai2018}\end{tabular}     & no      & non-structured    & yes  \\ 
  \begin{tabular}{@{}l@{}}\texttt{tf.contrib.model\_pruning} \\\quad\citep{Zhu2017}\end{tabular}  & no      & non-structured    & no  \\ 
  \begin{tabular}{@{}l@{}}RNN Pruning \\\quad\citep{Narang2017}\end{tabular}                      & no      & non-structured    & no  \\ 
  \begin{tabular}{@{}l@{}}Deep Compression\\\quad\citep{Han2015}\end{tabular}                     & no      & non-structured    & no  \\ 

  \midrule \midrule
  \begin{tabular}{@{}l@{}}Group-wise Brain Damage \\\quad\citep{Lebedev2015}\end{tabular}         & no      & channel           & no  \\ 
  \begin{tabular}{@{}l@{}}$L^1$-norm Channel Pruning \\\quad\citep{Li2016}\end{tabular}           & no      & channel           & no  \\ 
  \begin{tabular}{@{}l@{}}Structured Sparsity Learning (SSL)\\\quad\citep{Wen2016}\end{tabular}   & no      & channel/kernel/layer           & yes  \\ 
  \begin{tabular}{@{}l@{}}ThiNet\\\quad\citep{Luo2017}\end{tabular}                               & no      & channel           & no  \\ 
  \begin{tabular}{@{}l@{}}LASSO-regression Channel Pruning \\\quad\citep{He2017}\end{tabular}     & no      & channel           & no  \\ 
  \begin{tabular}{@{}l@{}}Network Slimming\\\quad\citep{Liu2017}\end{tabular}                     & no      & channel           & yes  \\ 
  \begin{tabular}{@{}l@{}}Sparse Structure Selection (SSS) \\\quad\citep{Huang2017}\end{tabular}  & no      & layer             & yes  \\ 
  \begin{tabular}{@{}l@{}}Principal Filter Analysis (PFA) \\\quad\citep{Suau2018}\end{tabular}    & no      & channel           & yes/no  \\ 

  \bottomrule
\end{tabular} 
\captionsetup{singlelinecheck=off,font=footnotesize,width=0.88\textwidth}
\caption*{
  We provide examples of different categories of methods.  
  This is not a complete list of methods.
}
\end{table*}

\paragraph{Strict parameter budget throughout training and inference}
This feature was discussed in depth in the main text.  
Most of the methods to date are \emph{compression} techniques, i.e. they start training with a fully parameterized, dense model, and then reduce parameter counts. 
To the best of our knowledge, only three methods, namely DeepR~\citep{Bellec2017}, SET~\citep{Mocanu2018} and ours, \emph{strictly} impose, throughout the entire course of training, a fixed small parameter budget, one that is equal to the size of the final sparse model for inference.  
We make a distinction between these \emph{direct training} methods (first block) and \emph{compression} methods (second and third blocks of Table~\ref{tb:tax}) \footnote{
  Note that an intermediate case is NeST~\citep{Dai2017,Dai2018}, which starts training with a small network, grows it to a large size, and finally prunes it down again. Thus, a fixed parameter footprint is not strictly imposed throughout training, so we list NeST in the second block of Table~\ref{tb:tax}.
}.

This distinction is meaningful in two ways:  
(a) practically, \emph{direct training} methods are more memory-efficient on appropriate computing substrate by requiring parameter storage of no more than the final compressed model size;  
(b) theoretically, these methods, if performing on par with or better than \emph{compression} methods (as this work suggests), shed light on an important question: whether gross overparameterization during training is necessary for good generalization performance?  

\paragraph{Granularity of sparsity}
The \emph{granularity} of sparsity refers to the additional structure imposed on the placement of the non-zero entries of a sparsified parameter tensor.  
The finest-grained case, namely \emph{non-structured}, allows each individual weight in a parameter tensor to be zero or non-zero independently.  
Early compression techniques, e.g.~\cite{Han2015}, and more recent pruning-based compression methods based thereon, e.g.~\cite{Zhu2017}, are non-structured (second block of Table~\ref{tb:tax}).  
So are all direct training methods like ours (first block of Table~\ref{tb:tax}).  

Non-structured sparsity can not be fully exploited by mainstream compute devices such as GPUs.
To tackle this problem, a class of compression methods, \emph{structured pruning} methods (third block in Table~\ref{tb:tax}), constrain ``sparsity'' to a much coarser granularity.  
Typically, pruning is performed at the level of an entire feature map, e.g. ThiNet~\citep{Luo2017}, whole layers, or even entire residual blocks~\citep{Huang2017}.  
This way, the compressed ``sparse'' model has essentially smaller and/or fewer \emph{dense} parameter tensors, and computation can thus be accelerated on GPUs the same way as dense neural networks. 

These \emph{structured compression} methods, however, did not make a useful baseline in this work, for the following reasons.
First, because they produce dense models, their relevance to our method (non-structured, non-compression) is far more remote than non-structured compression techniques yielding sparse models, for a meaningful comparison.  
Second, typical structured pruning methods substantially underperformed non-structured ones (see Table~2 in the main text for two examples, ThiNet and SSS), and emerging evidence has called into question the fundamental value of structured pruning: \cite{Mittal2018} found that the channel pruning criteria used in a number of state-of-the-art structured pruning methods performed no better than random channel elimination, and \cite{Liu2018} found that fine-tuning in a number of state-of-the-art pruning methods fared no better than direct training of a randomly initialized pruned model which, in the case of channel/layer pruning, is simply a less wide and/or less deep dense model (see Table~2 in the main text for comparison of ThiNet and SSS against \emph{thin dense}).  

In addition, we performed extra experiments in which we constrained our method to operate on networks with structured sparsity and obtained significantly worse results, see Appendix~\ref{app:granularity}. 

\paragraph{Predefined versus automatically discovered sparsity levels across layers}
The last key feature (rightmost column of Table~\ref{tb:tax}) for our classification of methods is whether the sparsity levels of different layers of the network is automatically discovered during training or predefined by manual configuration.  
The value of automatic sparsification, e.g. ours, is twofold.  
First, it is conceptually more general because parameter reallocation heuristics can be applied to diverse model architectures, whereas layer-specific configuration has to be cognizant of network architecture, and at times also of the task to learn. 
Second, it is practically more scalable because it obviates the need for manual configuration of layer-wise sparsity, keeping the overhead of hyperparameter tuning constant rather than scaling with model depth/size.  
In addition to efficiency, we also show in Appendix~\ref{app:control_allocation} extra experiments on how automatic parameter reallocation across layers contributed to its effectiveness.  

In conclusion, our method is unique in that it:
\begin{ls}
  \item strictly maintains a fixed parameter footprint throughout the entire course of training.
  \item automatically discovers layer-wise sparsity levels during training.  
\end{ls}

\section{Structured versus non-structured sparsity}
\label{app:granularity}

\begin{table*}[h]
\caption{
  Test accuracy\% (top-1, top-5) of Resnet-50 on Imagenet for different levels of granularity of sparsity. Numbers in square brackets are differences from the \emph{full dense} baseline.  
}
\label{tb:imagenet_granular}
\vspace{-3mm}\centering
\setlength\tabcolsep{5pt}
\begin{tabular}{ l | ll | ll }
  \toprule
  Final overall sparsity (\# Parameters) & \multicolumn{2}{c|}{$0.8$ (7.3M)}  & \multicolumn{2}{c}{$0.9$ (5.1M)} \\
  \midrule
  \emph{Thin dense}                
  & 72.4 {[-2.5]}     
  & 90.9 {[-1.5]}     
  & 70.7 {[-4.2]}     
  & 89.9 {[-2.5]}     \\ 
  \emph{Dynamic sparse (kernel granularity)}                
  & 72.6 {[-2.3]}      
  & 91.0 {[-1.4]}     
  & 70.2 {[-4.7]}     
  & 89.8 {[-2.6]}     \\ 
  \emph{Dynamic sparse (non-structured)}     
  & {\bf 73.3} {[{\bf -1.6}]} 
  & {\bf 92.4} {[{\bf \ 0.0}]} 
  & {\bf 71.6} {[{\bf -3.3}]} 
  & {\bf 90.5} {[{\bf -1.9}]} \\ 
  \bottomrule
\end{tabular} 
  
\end{table*}


We investigated how our method performs if it were constrained to training sparse models at a coarser granularity.  Consider a weight tensor of a convolution layer, of size $C_\text{out}\times C_\text{in} \times 3 \times 3$, where $C_\text{out}$ and $C_\text{in}$ are the number of output and input channels, respectively. 
Our method performed dynamic sparse reparameterization by pruning and reallocating individual weights of the 4-dimensional parameter tensor--the finest granularity.
To adapt our procedure to coarse-grain sparsity on groups of parameters, we modified our algorithm (Algorithm~1 in the main text) in the following ways:
\begin{enum}
  \item the pruning step now removed entire groups of weights by comparing their $L^1$-norms with the adaptive threshold.
  \item the adaptive threshold was updated based on the difference between the target number and the actual number of groups to prune/grow at each step.
  \item the growth step reallocated groups of weights within and across parameter tensors using the heuristic in Line 11 of Algorithm~\ref{alg}.
\end{enum}

\begin{figure}[t]
  \captionsetup{aboveskip=2pt,belowskip=0pt}
  \centering
    \includegraphics[width=0.45\textwidth]{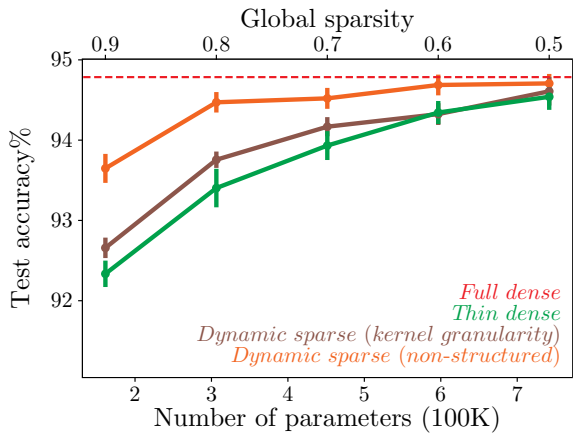}
  \caption{\footnotesize
    Test accuracy for WRN-28-2 trained on CIFAR10 for two variants of \emph{dynamic sparse}, i.e. kernel-level granularity of sparsity and non-structured (same as \emph{dynamic sparse} in the main text), as well as the \emph{thin dense} baseline.
    Conventions same as in Figure~\ref{fig:mnist_accuracy}. 
  }
  \label{fig:cifar_granular}
\end{figure}

We show results at kernel-level granularity (i.e. groups are $3\times3$ kernels) in Figure~\ref{fig:cifar_granular} and Table~\ref{tb:imagenet_granular}, for  WRN-28-2 on CIFAR10 and Resnet-50 on Imagenet, respectively.
We observe that enforcing kernel-level sparsity leads to significantly worse accuracy compared to unstructured sparsity. For WRN-28-2, kernel-level parameter re-allocation still outperforms the \emph{thin dense} baseline, though the performance advantage disappears as the level of sparsity decreases. Note that the \emph{thin dense} baseline was always trained for double the number of epochs used to train the models with dynamic parameter re-allocation. 

When we further coarsened the granularity of sparsity to channel level (i.e. groups are $C_\text{in} \times3\times3$ slices that generate output feature maps), our method failed to produce performant models.

\section{Multi-layer perceptrons and training at extreme sparsity levels}
\label{app:control_allocation}

\begin{figure}[t]
  \centering
  \begin{subfigure}[b]{0.45\textwidth}
    \phantomsubcaption
    \includegraphics[width=\textwidth]{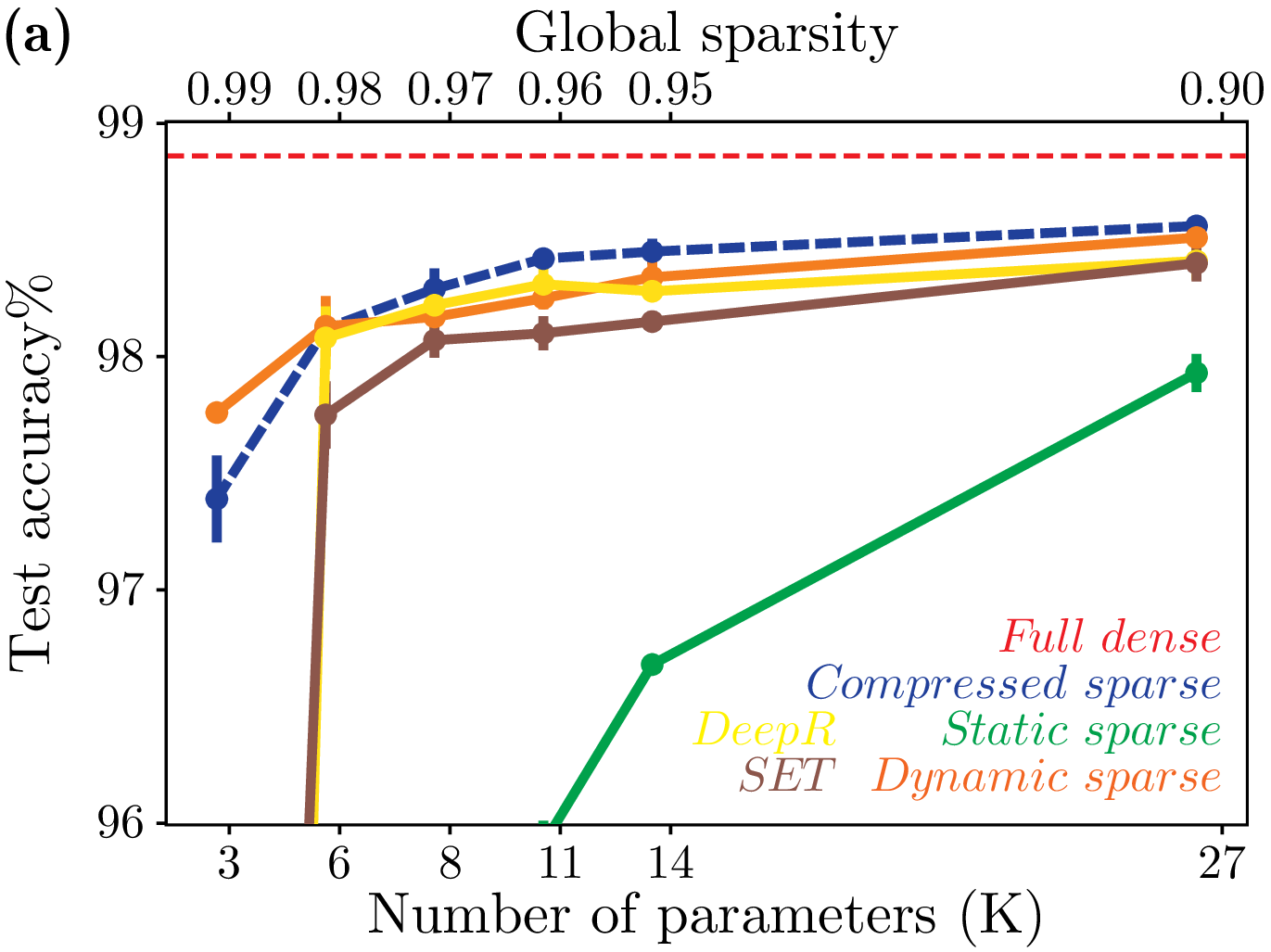} 
    \label{fig:mnist_accuracy}
  \end{subfigure}
  \quad
  \begin{subfigure}[b]{0.45\textwidth}
    \phantomsubcaption
    \includegraphics[width=\textwidth]{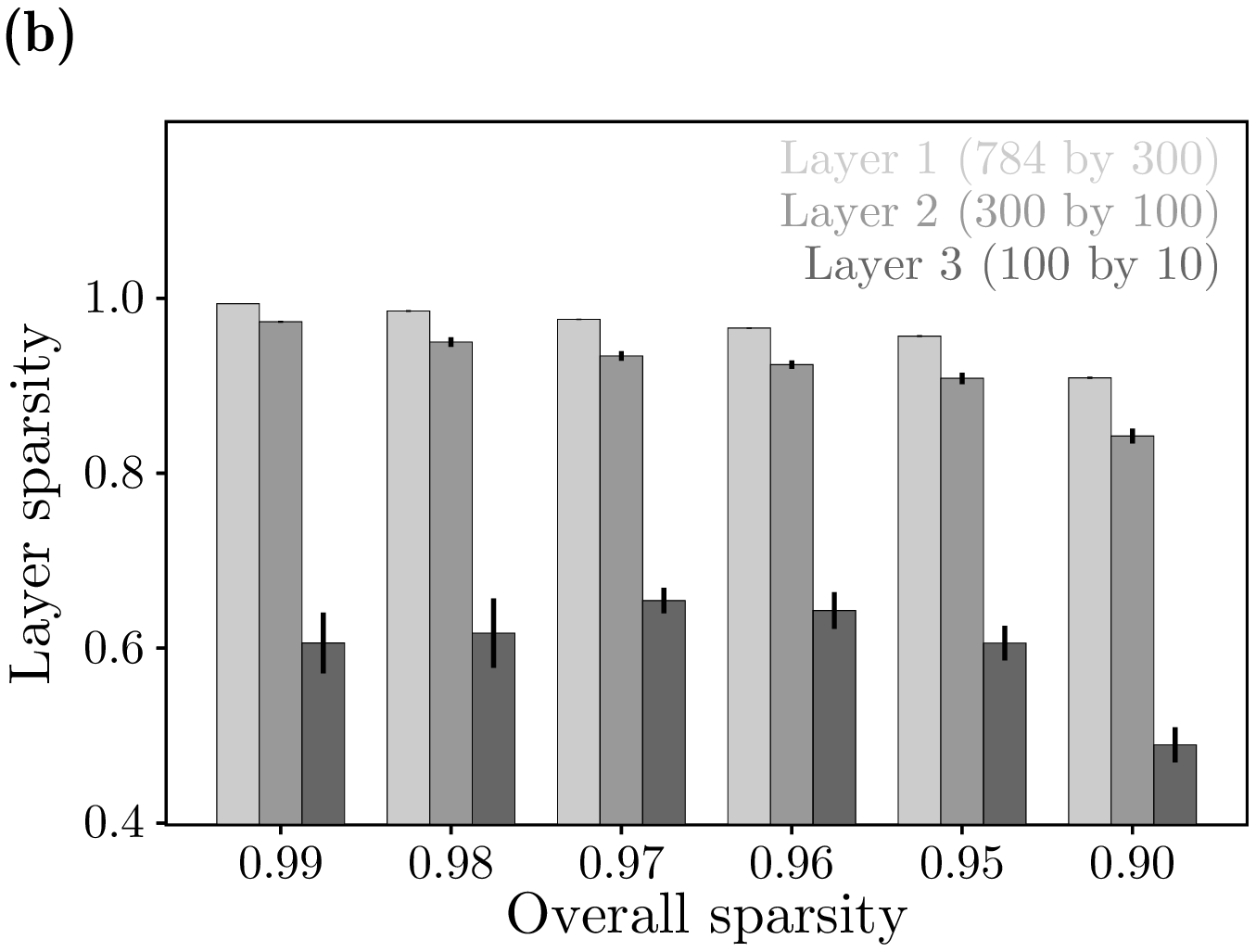} 
    \label{fig:mnist_sparsity}
  \end{subfigure}
  \caption{\footnotesize
    Test accuracy for LeNet-300-100-10 on MNIST for different training methods. Circular symbols mark the median of 5 runs, and error bars are the standard deviation. Parameter counts include all trainable parameters, i.e, parameters in sparse tensors plus all other dense tensors, such as those of batch normalization layers. Notice the failure of training at the highest sparsity level for \emph{static sparse}, SET, and DeepR.  
  }
  \label{fig:mnist_no_inter_layer_reallocation}
\end{figure}

We carried out experiments on small multi-layer perceptrons to assess whether our dynamic parameter re-allocation method can effectively distribute parameters in small networks at extreme sparsity levels. we experimented with a simple LeNet-300-100 trained on MNIST. Hyper-parameters for the experiments are reported in appendix~\ref{app:details}. The results are shown in Fig.~\ref{fig:mnist_accuracy}. Our method is the only method, other than pruning from a large dense model, that is capable of effectively training the network at the highest sparsity setting by automatically moving parameters between layers to realize layer sparsities that can be effectively trained. The per-layer sparsities discovered by our method are shown in Fig.~\ref{fig:mnist_sparsity}. Our method automatically leads to a top layer with much lower sparsity than the two hidden layers. Similar sparsity patterns were found through hand-tuning to improve the performance of DeepR~\cite{Bellec2017}. All layers were initialized at the same sparsity level (equal to the global sparsity level). While hand-tuning the per-layer sparsities should allow SET and DeepR to learn at the highest sparsity setting, our method automatically discovers the per-layer sparsities and allows us to dispense with such a tuning step.


\end{appendices}